\documentclass[10pt]{article} %
\usepackage[accepted]{tmlr}

\usepackage{amsmath,amsfonts,bm}

\def\eqref#1{equation~\ref{#1}}

\def\1{\bm{1}}

\DeclareMathAlphabet{\mathsfit}{\encodingdefault}{\sfdefault}{m}{sl}
\SetMathAlphabet{\mathsfit}{bold}{\encodingdefault}{\sfdefault}{bx}{n}

\newcommand{\R}{\mathbb{R}}

\DeclareMathOperator*{\argmax}{arg\,max}

\usepackage{hyperref}
\usepackage{url}

\renewcommand{\smallskip}[0]{\vspace{0.1cm}}

\newcommand{\BibTeX}{\rm B\kern-.05em{\sc i\kern-.025em b}\kern-.08em\TeX}

\def\X{{\mathcal X}}

\def\A{{\mathcal A}}
\def\S{{\mathcal S}}

\def\R{{\mathcal R}}

\def\Z{{\mathcal Z}}

\def\RR{{\mathbb{R}}}

\def\EE{{\mathbb{E}}}

\def\d{{\mathrm{d}}}

\usepackage{etoolbox}

\newtoggle{showold}

\togglefalse{showold} %

\definecolor{custom_green}{rgb}{0.1, 0.6, 0.5}
\definecolor{custom_red}{rgb}{0.9, 0.4, 0.35}
\definecolor{custom_blue}{rgb}{0.2, 0.45, 0.8}

\newlength{\paragraphxinterspace}
\newlength{\paragraphxabovespace}
\newcommand{\paragraphx}[2][\paragraphxabovespace]{%
  \par\vspace{#1}%
  \noindent\textbf{#2}\hspace{\paragraphxinterspace}\ignorespaces
}

\usepackage{float}
\usepackage{tabularx}
\usepackage{makecell}
\usepackage{adjustbox}
\usepackage[table]{xcolor}
\usepackage{booktabs}
\usepackage{tcolorbox}
\usepackage{bbm}
\usepackage{yhmath}
\usepackage{tablefootnote}

\usepackage{amsthm}
\usepackage{amsmath}
\usepackage{mathtools}

\usepackage{threeparttable}

\theoremstyle{plain}
\newtheorem{theorem}{Theorem}[section]

\theoremstyle{definition}

\newtheorem{example}[theorem]{Example}
\theoremstyle{remark}
\newtheorem{remark}[theorem]{Remark}

\definecolor{dirblue}{HTML}{B4DEFC}
\definecolor{compgreen}{HTML}{B2E6D2}
\definecolor{notavail_blue}{HTML}{DAE1E6}
\definecolor{notavail_blue2}{HTML}{EEF2F5}
\definecolor{dark_yellow}{HTML}{FFDE00}

\usepackage{tikz}
\usetikzlibrary{arrows.meta, trees, shapes.geometric}

\title{A Unified Framework for Zero-Shot Reinforcement Learning}

\author{\name Jacopo Di Ventura, Jan Felix Kleuker, Aske Plaat, Thomas Moerland \email \\ Leiden University}

\def\month{10}  %
\def\year{2026} %
\def\openreview{\url{https://openreview.net/forum?id=Nv26zx1Ywa}} %

\begin{document}

\maketitle

\begin{abstract}
Zero-shot reinforcement learning (RL) has emerged as a setting for developing general agents, capable of solving downstream tasks without additional training or planning at test-time. While conventional RL optimizes policies for fixed rewards, zero-shot RL requires learning representations that enable immediate adaptation to arbitrary reward functions. As the field matures, the growing diversity of approaches demands a foundational framework reconciling different perspectives under a common structure. In this work, we introduce a formal, unified framework for zero-shot RL, allowing for rigorous comparisons across methods. We propose a taxonomy organizing the algorithmic landscape along two levels: representation, distinguishing between compositional and direct methods based on their exploitation of value function decompositions; and learning paradigm, differentiating between reward-free and pseudo reward-free training. Additionally, we propose a unified view of existing error bounds, decomposing the total error into three primary contributing components: inference, reward, and approximation, serving as a foundation for more grounded comparisons of zero-shot methods.
\end{abstract}
\section{Introduction}
The reinforcement learning (RL; \citealt{sutton_reinforcement_2020}) objective is commonly defined as identifying a policy that maximizes the expected cumulative reward. While this paradigm has enabled remarkable progress across diverse domains \citep{berner2019dota,mirhoseini2021graph,mankowitz2023faster,seo2024avoiding}, it is limited by its dependence on a single reward function, hindering the ability to transfer to new objectives. %

Unsupervised RL relaxes this dependence by enabling agents to acquire environment knowledge in a task-agnostic pre-training phase, allowing for efficient fine-tuning once an extrinsic reward is revealed. Zero-shot RL \citep{touatidoes} takes this paradigm to its limit: agents must adapt to downstream tasks immediately after pre-training, requiring representations to be expressive enough for near-optimal behaviors to be extracted directly, without any task-specific fine-tuning, planning, or substantial computation \citep{touatidoes}. %
The zero-shot paradigm has emerged as a candidate for training behavioral foundation models, the RL analogue of foundation models \citep{tirinzonizero}; an area where RL has historically lagged behind, motivating further research in the field \citep{yang_foundation_2023}.
Recently, zero-shot RL has received increasing attention, resulting in a diverse set of proposed algorithms 
\citep{hansen_fast_2020,
liu2021aps,
frans2024unsupervised,
agarwal2025proto,
cetin2025finer,
sun2025unsupervised}, covering a wide range of algorithmic principles and design choices.

In this work, we establish the first unified framework for the field, introducing a taxonomy that groups methods according to their shared principles and underlying differences. This taxonomy, together with a consistent notation, provides the structure needed to navigate the zero-shot landscape. Note that our framework focuses on zero-shot-specific algorithmic paradigms and not, for example, on representation learning strategies within a given zero-shot paradigm. Each class in our framework is illustrated with a small set of representative examples from the literature. As such, our intention is to add structure to the field, not to fully survey it. We further introduce a unifying lens under which existing error bounds on zero-shot methods can be understood, decomposing the total error into three primary contributing factors.

\section{Preliminaries}
Formally, the reinforcement learning problem is modeled as a Markov Decision Process (MDP) \citep{stratonovich1960, puterman1994markov}, defined as a tuple $\mathcal{M}=(\mathcal{S}, \mathcal{A}, p, p_0, r, \gamma)$. Here, $\S$ is the state space, $\A$ is the action space, $p_0 \in \Delta(\S)$ is the initial state distribution, $p(s'|s,a) \in \Delta(\S)$ is the transition function,  $r(s,a,s')$ is the bounded reward function (see Appendix \ref{sec::app-Notation} for details of the reduction of arguments)
and $\gamma \in [0,1)$ is the discount factor. 
The behavior of an agent interacting with this MDP is defined by a policy $\pi\in\Pi$, that is, a mapping $\pi: \S \to \Delta(\A)$ assigning a distribution over actions to each state. The value function $V^\pi(s)$ denotes the expected discounted return starting from state $s$ and following policy $\pi$, $V^\pi(s) = \EE_{\pi,p}[\sum_{t=0}^\infty \gamma^t \,r(s_t,a_t)|s_0=s]$, resulting in the RL objective $\pi^\ast(\cdot| s) = \arg\max_{\pi \in \Pi} V^\pi(s)$.

\phantomsection
\label{sec:sr}
A core concept in unsupervised RL is that of occupancy measures. The successor representation (SR;~\citealt{dayan1993improving}) is simplest formulation of this concept. Formally, in a discrete MDP, for a given policy $\pi$, the SR is defined as the expected discounted future occupancy of state $s^+$ starting from state $s$:
\begin{equation}
\label{eq::sr}
    M^\pi(s, s^+) = \mathbb{E}_{\pi,p} \Big[ \sum\nolimits_{t=0}^{\infty} \gamma^t \mathbbm{1}\{s_t = s^+\} \mid s_0 = s \Big].
\end{equation}
This representation allows for a linear decoupling of the MDP dynamics from the reward function, where the value function is recovered as $ V^\pi(s) = \sum\nolimits_{s^+ \in \mathcal{S}} M^\pi(s, s^+) r(s^+),$
providing a theoretical basis for zero-shot evaluation of a policy under varying rewards.%

\paragraphx[-0.1em]{Zero-Shot RL}
In the zero-shot setting, we consider a family of MDPs, $\mathcal{M}^\mathcal{R} \equiv \left\{ (\mathcal{S}, \mathcal{A}, p, p_0, r, \gamma) \,\middle|\, r \in \mathcal{R} \right\}$, where $\mathcal{R}$ denotes a specified set of reward functions under consideration, which may represent the space of all possible rewards or just a subset of tasks or skills of interest \citep{barreto_successor_2017}. Under this setting, training is decoupled from task-specific feedback: while the agent may observe rewards during training, these signals are arbitrary and non-informative of the downstream tasks. At inference, reward functions are drawn from a distribution of downstream tasks $\mathcal{D}^\mathrm{test}$, which is unknown during training. The objective is to obtain optimal policies $\pi^*_r$ for all $r \in \mathcal{D}^\mathrm{test}$ without additional parameter optimization (fine-tuning), explicit planning (i.e. reasoning over state transitions to synthesize new behaviors), or substantial computation. Fine-tuning and planning are prohibited as both produce policies not already encoded in the learned representation: fine-tuning by drawing on additional test-time data, planning by synthesizing new behaviors through an auxiliary structure such as a world model, contradicting the premise of zero-shot adaptation. The third constraint, computation itself, is formalized in Section~\ref{sec:zero_shot_boundary} through the notion of instant extraction. %
The objective for zero-shot RL can formally be expressed as

\begin{equation}
    \label{eq::donwstream_test}
    \pi^* = \arg\max_{\pi \in \Pi}\EE_{r\sim \mathcal{D}^\mathrm{test}}[ \EE_{p_0,\pi_r(\cdot|s),p}[\sum\nolimits_{t=0}^\infty \gamma^t\,r(s_t,a_t,s_{t+1})]].
\end{equation}

\section{Framework}

\begin{figure}[htbp]
    \begin{center}
        \usetikzlibrary{calc}

\begin{tikzpicture}
[
    level 1/.style={sibling distance=6.5cm, level distance=20pt},
    level 2/.style={sibling distance=4.1cm, level distance=35pt},
    level 3/.style={sibling distance=2cm, level distance=35pt},
    edge from parent/.style={draw, -{Stealth[length=2mm]}, semithick},
    every node/.style={font=\footnotesize, inner sep=2pt},
    base/.style={rectangle, rounded corners=3pt, draw=gray!20, fill=gray!5, inner sep=4pt, align=center},
    primary/.style={base, fill=compgreen, draw=white!0},
    secondary/.style={base, fill=dirblue, draw=white!0},
    label_node/.style={font=\scriptsize\itshape, text=black!70},
    stroke_green/.style={draw=compgreen!160},
    stroke_blue/.style={draw=dirblue!160},
    blue_unreachable/.style={fill=notavail_blue2, draw=notavail_blue, minimum width=38pt,minimum height=16pt},
    forked/.style 2 args={
        xshift=#2,
        edge from parent path={
            (\tikzparentnode.south) -- +(0,-#1) -| (\tikzchildnode.north)
        }
    },
    leaf_style/.style={font=\scriptsize, text=black!70,
                        rounded corners=3pt,align=center, inner sep=4pt, fill=gray!5},
    comp_free_leaf/.style={draw=compgreen!160,fill=compgreen!20},
    comp_pseudo_leaf/.style={dashed, draw=compgreen!160, fill=compgreen!20},
    dir_pseudo_leaf/.style={dashed, draw=dirblue!160, fill=dirblue!20},
    top_style/.style={draw=dark_yellow},
    modern_dash/.style={
    line width=0.8pt,
    dash pattern=on 2.5pt off 1.5pt
    },
    legend_text/.style={font=\scriptsize\bfseries, text=black!70, inner sep=2pt},
    legend_line/.style={black!70, thin},
]
\node[base, thick, top_style] (root) {$Q_r^\ast = \mathcal F(\mu,r)$}

    child {
        node[secondary] {$\mathcal F = \mathrm{Id}_\mu$}
        child {
            node[secondary, stroke_blue, modern_dash] {$\mu^r(s,a)$}
            child[yshift=5pt] { node[leaf_style,dir_pseudo_leaf, modern_dash] {HILP (\ref{sec::hilbert})} }
            child[yshift=5pt] { node[leaf_style,dir_pseudo_leaf, modern_dash] {FRE (\ref{sec:fre})} }
            edge from parent[] node[left, label_node, yshift=3pt, xshift=-4pt] {\shortstack{pseudo\\reward-free}}
        }
        child {
            node[secondary, blue_unreachable] {$-$}
            edge from parent node[right, label_node,  xshift=0pt, yshift=3pt] {reward-free}
        }
        edge from parent[dirblue] node[right, label_node, yshift=5pt, xshift=-16pt] {direct}
    }
    child {
        node[primary] {$\mathcal F \neq \mathrm{Id}_\mu$}
        child {
            node[primary, stroke_green, modern_dash] {$\mu^r(s,a)$}
            child[yshift=5pt] { node[leaf_style,comp_pseudo_leaf, modern_dash] {\shortstack{USF (\ref{sec:usf})}} }
            child[yshift=5pt] { node[leaf_style,comp_pseudo_leaf, modern_dash] {\shortstack{FB (\ref{sec::FB})}} }
            edge from parent[] node[left, label_node, yshift=6pt] {\shortstack{pseudo\\reward-free}}
        }
        child {
            node[primary, stroke_green] {$\mu^\pi(s,a)$}
            child[yshift=5pt, xshift=-3pt] { node[leaf_style,comp_free_leaf] {\shortstack{SF\&GPI (\ref{sec::SF})}} }
            child[yshift=5pt, xshift=0pt] { node[leaf_style,comp_free_leaf] {\shortstack{PSM (\ref{sec::psm})}} }
            edge from parent node[right, label_node, yshift=3pt] {reward-free}
        }
        edge from parent[compgreen] node[left, label_node, yshift=5pt, xshift=39pt] {compositional}
    };

    \node[legend_text, anchor=east] at ($(root.west)+(-4.8cm,-3pt)$)  {Representation};
    \node[legend_text, anchor=east] at ($(root.west)+(-5.6cm,-34pt)$) {\shortstack{Learning\\Paradigm}};

    \end{tikzpicture}
    \end{center}
    \caption{Taxonomy of zero-shot RL methods, organized along two axes that jointly determine how each method recovers the optimal action-value function $Q_r^\ast = \mathcal{F}(\mu, r)$ from a learned representation $\mu$ and a reward function $r$. The first split (\textbf{representation}) separates \textcolor{dirblue!200}{direct} methods from \textcolor{compgreen!200}{compositional} methods: direct methods define $\mu$ as the value function itself $\mu=Q^*_r$, so the decomposition operator reduces to the identity, $\mathcal{F} = \mathrm{Id}_\mu$; compositional methods instead define $\mu$ as some intermediate representation that requires a non-trivial operator, $\mathcal{F} \neq \mathrm{Id}_\mu$, to reconstruct task-specific value functions. The second split (\textbf{learning paradigm}) separates pseudo reward-free training (dashed boxes), where the target $\mu^r(s,a)$ is conditioned on the reward function, from reward-free training (solid boxes), where the target $\mu^\pi(s,a)$ is learned independently of any reward. Because direct representations are reward-conditioned by construction, a reward-free direct method is structurally infeasible, marked by the grayed-out node ($-$).
    }
    \label{fig:01}
\end{figure}

Our taxonomy categorizes the space of zero-shot methods by identifying two primary decision nodes: representation and learning paradigm. At the first level, we distinguish between methods based on their representation, partitioned into direct and compositional approaches. The taxonomy further divides based on the learning paradigm: reward-free vs. pseudo reward-free. This hierarchical decomposition, illustrated in Figure \ref{fig:01}, serves as the basis for our framework. The following section provides a breakdown and formal definitions of these design primitives.

\subsection{Representation: Direct vs. Compositional}

Direct methods provide a conceptually straightforward approach to zero-shot RL by learning a reward-conditioned value function $Q_r(s,a)$, also known as a universal value function \citep{schaul2015universal}. In contrast, compositional methods decompose the value function by learning an intermediate target, enabling the reconstruction of task-specific value functions during inference.

\paragraphx[-0.1em]{Direct representations} learn a direct mapping from state-action and reward to optimal values:
\begin{equation}
    \label{eq::UVF}
    Q^*:\mathcal S\times\mathcal A\times\mathcal R\to\mathbb R, \quad (s,a,r) \mapsto Q^*_r(s,a).
\end{equation}
Under this direct mapping, the function approximator must capture the complete reward-induced structure, yielding $Q^*$ values directly from $(s,a,r)$, with no explicit substructure between policy, occupancy, and value. Policy extraction for a target task $r \sim \mathcal{D}^\mathrm{test}$ is performed via $\pi^*_r(s) \in \arg\max_a Q^*(s,a,r)$.

\paragraphx[-0.1em]{Compositional representations} explicitly leverage the structure of the value function. Under this approach, the learning problem reduces to estimating individual components, that are recombined at inference time according to the specified substructure. More formally, compositional methods learn representations $\mu(s,a)$, which allow to infer optimal values via some structure in the form of a decomposition operator $\mathcal{F}$:
\begin{equation}
    \label{eq::decomp-methods}
    Q^*_r(s,a) = \mathcal{F}(\mu, r),
\end{equation}
where the operator $\mathcal{F}$ encodes the relationship between the learned representation and the reward. An instantiation of $\mathcal{F}$ could be as simple as the inner product between some reward representation $f(r)$ and the target, i.e. $\mathcal{F}(\mu,r)= \langle \mu, f(r) \rangle$, as for the successor representation (cf. Eq. \ref{eq::sr}).

Notably, as shown in Figure~\ref{fig:01}, direct representations constitute the special case where $\mu^r = Q_r^\ast$, or equivalently $\mathcal F = \mathrm{Id}_\mu$. Because we define compositional methods to be everything that is not direct, we assume compositional methods to have a non-trivial decomposition operator $\mathcal F \neq \mathrm{Id}_\mu$.

\subsection{Learning Paradigm: Reward-free vs. Pseudo reward-free}
At test time, the agent is evaluated under an unknown reward distribution $\mathcal{D}^{\mathrm{test}}$, to which it has no access during training. Consequently, it must either adopt a reward-free training paradigm or define an arbitrary set of reward functions to guide the learning process (pseudo reward-free).

\paragraphx[-0.1em]{Reward-free} methods learn a policy-conditioned quantity $\mu^\pi(s,a)$ -- always superscribed by $\pi$ -- through objectives that are entirely independent of reward signals. By definition, this paradigm excludes any objective relying on reward maximization (i.e. Bellman optimality backups). Therefore, as illustrated in Figure \ref{fig:01}, reward-free objectives only exist in the context of compositional representations.

As an example, consider a discrete-space MDP and the successor representation of a fixed policy $\pi$, denoted as $M^\pi$, learned by minimizing the TD-error of the SR following $\pi$. The representation is compositional because it targets a component of the value function; as such, we denote it as $\mu^\pi = M^\pi$,  with superscript $\pi$ because it is learned via a reward-free objective. Assuming that the policy space $\Pi$ is large enough, that is, containing the optimal policy for all $r \in \R$ Zero-shot adaptation to any reward $r \in \mathcal{R}$, is performed via the operator $\mathcal{F}$ as $Q^\ast_r = \mathcal{F}(\mu^\pi, r) = \max_{\pi\in\Pi}  \langle M^\pi, r \rangle$.

\paragraphx[-0.1em]{Pseudo reward-free} methods, in contrast, leverage reward signals to learn representations contingent on a reward function, denoted by $\mu^r(s, a)$. The agent is provided access to a training distribution of random rewards, $r \sim \mathcal{D}^{\text{train}}$, which are non-informative of the downstream tasks, resulting in a form of self-supervised RL. The underlying principle is that by sampling a sufficiently diverse set of random reward functions during training, the resulting representation space will cover the rewards encountered during inference.

\vspace{5pt}
Finally, we formalize the preceding concepts into a unified zero-shot RL framework.
\label{box::zero_shot_framework}
\begin{tcolorbox}
[
    title=Unified Zero-Shot RL Framework,
    toptitle=0.5mm,
    center,
    width=0.8\textwidth,
    bottomtitle=0.5mm,
    colback=gray!5!white,
    colframe=gray!90!black,
    boxrule=0.5pt
]
    \textbf{Training}\\
    Define $\mu(s,a)$ and a compatible learning paradigm such that
    \begin{equation}
        Q_r^* = \mathcal{F}(\mu,r) \quad \forall r \in \R
    \end{equation}
    \textbf{Policy Extraction}\\
    Given a (unseen) reward $r \sim \mathcal{D}^\mathrm{test}$ extract a policy via 
    \begin{equation}
        \pi^*_r(\cdot|s) = \arg\max_a \mathcal{F}(\mu,r)
    \end{equation}
\end{tcolorbox}%

\subsection{The Zero-shot Boundary}
\label{sec:zero_shot_boundary}

Our framework exposes an ambiguity in the definition of zero-shot RL: how much computation is the decomposition operator $\mathcal F$ allowed at test time? Fine-tuning $\mu$ and explicit planning are clearly ruled out, but beyond the vague prohibition on ``substantial computation'' no upper bound on the cost of $\mathcal F$ has been specified. We propose the notion of instant extraction as a reference point. An operator $\mathcal F$ admits instant extraction if its cost is $O(1)$ in $|\Pi|$, that is, if extraction does not scale with the number of behaviors encoded in $\mu$. Pseudo reward-free methods satisfy this: the policy is obtained by a single conditioning operation on the desired reward, independently of what $\mu$ encodes. This is zero-shot RL in the strictest sense. However, not every method commonly accepted as zero-shot meets this criterion. A test-time search over the policies already encoded in $\mu$ has cost growing with $|\Pi|$ and so fails instant extraction, yet it only selects among represented behaviors rather than synthesizing new ones, and completes in a modest number of operations under mild assumptions. It is unclear at what point such a search ceases to be zero-shot. Fixing a universal bound on $\mathcal F$ is impractical given the diversity of extraction procedures. We therefore propose to treat instant extraction as the strict end of the zero-shot definition and to position each method's $\mathcal F$ relative to it. Methods that admit instant extraction are unambiguously zero-shot; beyond that lies a fuzzy region, and a method's distance from instant extraction gives an axis along which to place it.

In the following sections we review key zero-shot methods starting with direct methods in Section~\ref{sec:direct}, and moving to compositional in Section~\ref{sec:comp}. In Section \ref{sec:error_decomp}, we analyze error bounds for different algorithms under a common lens, categorizing the total error into three primary components showing how design choices affect the error. An extensive list of zero-shot methods categorized under our taxonomy can be found in Table \ref{table:feature_matrix_table}.

\section{Direct Representations}
\label{sec:direct}

Methods relying on direct representations (left branch in Figure~\ref{fig:01}) parameterize and optimize the value function directly. The training objective is the optimal value function itself, $\mu^r(s,a) = Q^*_r(s,a)$. %
While some methods instead learn a parameterized policy $\pi^*_r(s)\in \Delta(\mathcal{A})$, this often serves as a practical workaround for the intractability of $\arg\max_a Q^*_r(s,a)$ ~\citep{Lillicrap2016DDPG, Fujimoto2018TD3}. Because we are only concerned with \emph{optimal} behavior for each task, direct representations implicitly parameterize each policy by the reward function for which it is optimal. This approach is usually facilitated via an embedding function $f: \mathcal{R} \to \mathcal{Z}$ that maps reward functions into a latent representation space. This allows us to index the set of optimal policies by $\mathcal{Z}$, where $\pi_z$ represents the optimal policy for the reward function embedded as $z = f(r)$.

\paragraphx[-0.1em]{Defining a task space} While $f$ could theoretically be the identity mapping, this becomes intractable in continuous or high-dimensional settings. Consequently, training direct representations generally involves two phases: (i) define or learn a task encoder $f: \R \mapsto \mathcal{Z}$;
(ii) train some representation of optimal action-value functions or policies  $\mu^{f(r)}$ using rewards $r$ sampled from some distribution until convergence. Since the second phase corresponds to standard RL training for each sampled reward, the primary challenge lies in the first phase: learning an encoder $f$ that defines a task space $\mathcal{Z}$ expressive enough to retain all reward-relevant information while remaining sufficiently smooth to enable generalization through $\mu$. Since the space of admissible rewards is vast, the effectiveness of generalization critically depends on how the chosen prior constrains and shapes coverage of plausible downstream tasks. In addition, task embeddings suffer from inherent identifiability issues: distinct reward functions may yield indistinguishable behaviors, making the corresponding task representations fundamentally ambiguous. In the following, we present three methods that successfully define or learn such a space.
\vspace{-0.2em}
\subsection{Goal-Conditioned RL}
A canonical example of such an embedding arises in goal-conditioned RL (GCRL;~\citealt{schaul2015universal, andrychowicz2017hindsight, levy2018hierarchical, chebotar2021actionable, eysenbach2022contrastive, park2023hiql}), where tasks are limited to goal reaching. In this setting, the reward embedding emerges naturally: each task is defined by a goal state, giving rise to the natural identification $\mathcal{Z} = \mathcal{S}$. %
\vspace{-0.2em}
\subsection{Hilbert Representations}\label{sec::hilbert}

The Hilbert Representations framework (HILP;~\citealt{park2024foundation}) defines a geometric abstraction of the state space by learning a mapping $\phi:\mathcal{S} \to \mathcal{Z}$, where $\mathcal{Z}$ is a Hilbert space, such that distances in the latent space correspond to temporal distances in the original MDP. Using this, we can define a goal-reaching reward function for any vector in latent space via $r_z(s,s') \coloneqq (\phi(s')-\phi(s))^\top z$, rewarding transitions that align with the direction of the goal $z$. Through $r_z(s,s')$, optimal policies $\pi^*_z$ can be trained for all $z \in \mathcal{Z}$. At test time, given a reward function $r$, the optimal behavior for any task can be recovered by $\pi_{f(r)}$ for $f(r) = \arg\min_{z\in \mathcal Z}[(r(s,a,s')-r_z(s,s'))^2]$.

\vspace{-0.2em}
\subsection{Functional Reward Encoding}\label{sec:fre}

Restricting the task space to goal-reaching is not intrinsic to direct representations. The Functional Reward Encoding framework~\citep[FRE;][]{frans2024unsupervised} introduces a method to embed arbitrary reward functions into a latent space $f_e: \R \to \mathcal{Z}$. A state-dependent reward function $r:\mathcal{S}\to\mathbb{R}$ is represented by a set of samples $\{(s_i, r(s_i))\}_{i \in \mathcal{I}}$, which a permutation-invariant transformer encodes into a latent vector $z \in \mathcal{Z}$. To ensure $z$ captures generalizable features of the reward landscape, an encoder-decoder architecture is trained using an information-bottleneck objective over a reward distribution $\mathcal{D^{\text{train}}}$ comprised of goal-reaching, linear, and random MLP functions. %
Then, networks $Q(\cdot, z)$, $V(\cdot, z)$, and $\pi(\cdot, z)$ are trained using Implicit Q-Learning~\citep{kostrikov2022offline} under sampled reward functions projected into the latent space $\mathcal{Z}$ via the frozen encoder.

\paragraphx[-0.1em]{Relation to Framework} For all direct methods, the training target is the optimal action-value function itself, i.e. $\mu = Q_r^\ast$. In general, $\mu$ cannot be conditioned directly on the reward function, as the reward space is generally intractable; instead, a suitable reward representation $f : \R  \rightarrow \Z$ is required such that $\mu^{f(r)} = Q_r^\ast$. Given a reward function $r$, a policy can then be extracted as
$
\pi_r(\cdot|s) = \arg\max_a \mu^{f(r)}(s,a).
$

\section{Compositional Representations}\label{sec:comp}

Compositional methods (right branch in Figure~\ref{fig:01}) are built on occupancy measures such as the SR (Eq.~\ref{eq::sr}) \citep{agarwal2025unified}. They aim to capture the occupancy measure of a set of policies, including the degenerate case of a single policy, and to extract relevant behavior given a reward, either through an explicit search at test time or by shifting this cost to pretraining via inductive biases. Such methods can be reward-free (Sections~\ref{sec:sf_GPI}, \ref{sec::psm}) or pseudo reward-free (Sections~\ref{sec:usf}, \ref{sec::FB}). This section covers the key zero-shot methods together with their foundational objects (Sections~\ref{sec::SF}, \ref{sec:sm}), organized by conceptual evolution rather than training paradigm: starting from basic approaches, we trace how subsequent methods address specific limitations or pursue distinct directions.

\subsection{Successor Features}\label{sec::SF}

Successor Features (SF; \citealt{barreto_successor_2017}) extend the SR to continuous MDPs by tracking expected discounted future \emph{features} instead of states. Assume a feature mapping $\phi: \mathcal{S} \times \mathcal{A} \times \mathcal{S} \to \mathbb{R}^d$ for some $d \in \mathbb{N}$, and suppose that the reward function can be linearly decomposed as
\begin{equation}
    \label{eq::sf_reward_factorization}
    r(s,a,s') = \phi(s,a,s')^\top w,
\end{equation}
where $w \in \mathbb{R}^d$ is a weight vector and $\phi(s,a,s') \in \mathbb{R}^d$ are the \textit{basic features} of the one-step transition $(s,a,s')$. Note, that for a single reward function, such a decomposition always exists (e.g., $\phi(s,a,s') = r(s,a,s')$ and $w = 1$). Following Eq.~\ref{eq::sr}, the successor features of a policy $\pi$ are defined as
\begin{equation}
    \label{eq::sf}
    \psi^\pi(s,a) \coloneqq 
    \mathbb{E}_{\pi,p} \left[
        \sum\nolimits_{t=0}^{\infty} \gamma^t 
        \phi(s_t,a_t,s_{t+1})
        \,\Big|\, s_0 = s, a_0 = a
    \right].
\end{equation}

Just as the SR decouples dynamics from rewards, SFs decouple dynamics from the reward weights $w$. Given a reward embedding $w$ and basic features $\phi$, the action-value function under policy $\pi$ can be recovered as 
$
Q^\pi_r(s,a) = \psi^\pi(s,a)^\top w.
$
This decomposition allows for zero-shot policy evaluation: the performance of $\pi$ can be evaluated on any task in the span of $\phi$ by simply updating $w$.

\paragraphx[-0.1em]{Training} Analogous to value functions, successor features satisfy a Bellman-like recursion: 
\begin{equation}
\label{eq::td-error-sf}
    \psi^\pi(s,a) = \mathbb{E}_{s'\sim p,\, a'\sim\pi}[\phi(s,a,s') + \gamma \psi^\pi(s', a')].
\end{equation}
Allowing $\psi^\pi$ to be learned using standard temporal-difference (TD) methods.

While the linearity assumption of Eq.~\ref{eq::sf_reward_factorization} may appear restrictive, the framework can still be applied when $\phi$ only \emph{approximately} linearizes rewards by considering the best linear approximation $\tilde{w}_r$ such that $r \approx \phi^\top \tilde{w}_r$.  
For instance, $\tilde{w}_r$ can be obtained by minimizing the mean squared error:
\begin{equation}
    \label{eq::lin_reg_task_sf}
    \tilde{w}_r = 
    \arg\min_w \,
    \mathbb{E}_{(s,a,s')\sim \mathcal{D}}\big[
        (r(s,a,s') - \phi(s,a,s')^\top w)^2
    \big].
\end{equation}

\phantomsection\label{sec:sf_GPI}
\paragraphx[-0.1em]{Generalized Policy Improvement} Standard policy improvement \citep{sutton_reinforcement_2020} identifies a better policy relative to the value function of a single policy, Generalized Policy Improvement (GPI; \citealt{barreto_successor_2017}) extends this logic to a set of policies $\{\pi_i\}_{i=1}^n$. Given the respective action-value functions $\{Q^{\pi_i}\}_{i=1}^n$ under the same reward, we define the GPI policy as the point-wise maximum over the set
\begin{equation}
   \pi_{\text{GPI}}(\cdot| s) \in \arg\max_a \max_i Q^{\pi_i}(s, a).
\end{equation}
Similarly to the standard policy improvement theorem, it can be shown the new policy guarantees that $Q^{\pi_{\text{GPI}}}(s, a) \geq \max_i Q^{\pi_i}(s, a)$ for all $s \in \mathcal{S}$ and $a \in \mathcal{A}$.

\paragraphx[-0.1em]{Relation to Framework} Pairing SFs with GPI is an instance of a reward-free approach, as the training target $\mu^\pi=\psi^\pi$ can be trained without any reward signal. From said target, the optimal value function can be recovered using the decomposition operator $\mathcal{F} (\mu^\pi,r) = \max_\pi (\mu^\pi)^ w_r$. In the policy extraction phase, given a reward function $r \sim \mathcal{D}^\mathrm{test}$, we first calculate $\tilde{w}_r$ (Eq.~\ref{eq::lin_reg_task_sf}), for which we can infer an approximation for the optimal policy via GPI
\begin{equation}
     \pi_r(\cdot|s) = \arg\max_a \max_{i} \psi^{\pi_i}(s,a)^\top \tilde{w}_r.
\end{equation}

\subsection{Universal Successor Features}
\label{sec:usf}

The scalability of SF\&GPI is primarily constrained by the limited generalization across policy space. %
One might mitigate this issue by conditioning the successor features on the policy itself (i.e. learning $\psi^\pi(s, a)$ as a function of $\pi$), yet a naive search over $\Pi$ at test time remains computationally intensive without significant constraints. %
Universal Successor Features ~\citep[USF;][]{borsa2018universal} circumvent this limitation by exploiting the relationship between optimal policies and their inducing reward functions. By parameterizing policies by the reward weights for which they are optimal, we avoid an explicit search over policies during the policy extraction phase. Specifically, for a reward function of the form $r=\phi^\top w$, we define $\pi_w \coloneqq \pi^\ast_{\phi^\top w}$ as the optimal policy given the reward weights $w$. Then, the universal successor features for a family of policies $(\pi_w)_{w\in \mathcal W}$ are defined as
\begin{equation}
    \pi_{w}(\cdot | s)=\arg\max _a \psi(s,a,w)^\top w,\quad \psi(s,a,w) = \psi^{\pi_w}(s,a).
\end{equation} 
Notably, $\psi(s,a,w)$ encodes the successor features for the optimal policy with respect to all rewards admitting the factorization $r=\phi^\top w$. Therefore, the policy extraction phase for USF reduces to computing the task vector $w$ as in Eq.~\ref{eq::lin_reg_task_sf}.

\paragraphx[-0.1em]{Training} As with successor features, USF also satisfy a Bellman-like equation, motivating minimization of the Bellman residual to train USFs. Following Eq.~\ref{eq::td-error-sf}, we get:
\begin{equation}\label{eq::usf_td_loss}
        L(\psi) = \mathbb{E}_{w\sim p_\mathcal W}\!\left[\delta_{w}(t)^\top\, \delta_{w}(t) \right], \,\,
        \delta_{w}(t) = \phi(s_t, a_t, s_{t+1})+ \gamma\,\psi(s_{t+1},\pi_w(s_{t+1}), w)- \psi(s_t, a_t, w),
\end{equation}
where $\delta_w^t$ is the temporal-difference error~\citep{barreto_successor_2017}. 
These objectives are particularly useful when evaluating policies on tasks different from those that induced them, as in GPI.

\paragraphx[-0.1em]{The curse of $\phi$}
Successor Features rely on the assumption that all downstream tasks of interest lie in the linear span of the basic features $\phi$ (Eq.~\ref{eq::sf_reward_factorization}). Unless a unifying object encompassing both $\psi$ and $\phi$ is introduced, they cannot be learned jointly. In such case, two distinct objectives must be defined: one for $\psi$ and one for $\phi$. Choosing an appropriate training objective for learning $\phi$ is critical, as it determines which aspects of the environment are preserved and, consequently, which reward functions admit accurate linear parameterizations\footnote{For a comprehensive review of existing feature-learning methods for USF, we refer the reader to \citep{touatidoes}.}. Moreover, since $\phi$'s objective is defined independently of $\psi$, nothing guarantees that the resulting features are well-suited for accurately estimating successor features on top of them, potentially compounding error across the two stages.

TD-JEPA~\citep{bagatella2026tdjepa} narrows this gap by training the basic features $\phi$ and a predictor $T_\phi$ jointly via a single temporal-difference objective: for a family of policies $\pi_w(\cdot|s) = \arg\max_a T^\phi(\phi(s),a,w)^\top w$, defined as for USF above,
\begin{equation}
\label{eq::tdjepa_loss}
L(\phi, T^\phi) = \EE_{w\sim p_{\mathcal{W}}, a'\sim \pi_w(\cdot|s')}[\big\Vert T^\phi(\phi(s),a,w) - \bar\phi(s') - \gamma \, \bar T^\phi(\bar \phi(s'),a',w)\big\Vert^2],
\end{equation}
where $\bar T^\phi$ and $\bar \phi$ indicate target networks. Since this target mirrors the Bellman recursion satisfied by successor features (cf. Eq.~\ref{eq::td-error-sf}), $T_\phi$ is trained to approximate the successor features $\psi(\phi(s),a,w)$ of $\phi$. Crucially, $\phi$ appears in both the prediction and target of the loss, so it is learned jointly with $T_\phi$.%

\paragraphx[-0.1em]{Relation to Framework} The training target for USF-based methods is $\mu^r = \psi(s,a,w_r)$, allowing to recover the optimal values via the decomposition operator $\mathcal{F}(\mu^r,r) = (\mu^r)^\top w_r$. As such, we require some reward representations during training (cf. Eq. \ref{eq::usf_td_loss}), rendering this a pseudo reward-free approach. At inference time, given a reward $r$, we compute $\tilde{w}_r$ as Eq.~\ref{eq::lin_reg_task_sf} and infer a policy as $\pi_r(\cdot|s) = \arg\max_a \psi(s,a,\tilde{w}_r)^\top\tilde{w}_r$.

\subsection{Successor Measure}\label{sec:sm}
While feature-based methods adapt successor representations to continuous domains by tracking state properties, the successor measure (SM; \citealt{blier2021learning}) provides a more fundamental generalization. By transitioning from discrete counts to a measure-theoretic framework, it extends the SR to continuous state-action spaces while preserving its core interpretation. We overload the notation by identifying the SM as $M$, just as with the SR, as the SM naturally encompasses the SR.

Formally, the successor measure for a policy $\pi$, a state--action pair $(s,a) \in \mathcal{S}\times\mathcal{A}$, and a set $X \subset \mathcal{S}$ is defined as
\begin{equation}\label{eq::sm}
    M^\pi(s,a,X) \coloneqq \mathbb{E}_{\pi,p}[\sum\nolimits_{t=0}^\infty \gamma^t\,\mathbbm{1}\{(s_t,a_t)\in X\}\,|\, s_0=s, a_0=a].
\end{equation}

Treating $M^\pi$ as a measure allows us to handle discrete and continuous spaces in a unified way.  
Let $\rho \in \Delta(\mathcal{S})$ be a state distribution satisfying $\rho(s) > 0$ for all $s \in \mathcal{S}$, and define the Radon–Nikodym derivative $m^\pi \coloneqq \frac{\d M^\pi}{\d \rho}$, or in other words the density of $M^\pi$ with respect to $\rho$.  
For $\gamma < 1$, $M^\pi$ is a finite measure since $M^\pi(s,a,\mathcal{S}) = (1-\gamma)^{-1}$, which, together with the assumption on $\rho$, guarantees that this density exists.  We can then express the action–value function under a reward $r:\mathcal{S} \to \mathbb{R}$ as $Q^\pi(s,a) = \EE_{s^+ \sim \rho}[m^\pi(s,a,s^+)\, r(s^+)]$.

\paragraphx[-0.1em]{Training} Successor measures follow a Bellman-like relationship, allowing $m_\theta \approx m^\pi$ to be trained via temporal-difference learning, i.e.:
\begin{equation}
    \label{eq::SMloss}
    L(\theta) = 
    \EE_{\substack{s,a\sim\rho, s^+\sim\rho}}
    [(m_\theta(s,a,s^+) - \hat m_{\bar{\theta}}(s,a,s^+))^2],
\end{equation}
Where the target $\hat m_{\bar \theta}(s,a,s^+) = \frac{\delta_s(\d s^+)}{\rho(\d s^+)} + \gamma \EE_{s'+\sim p(\cdot|s,a), a'\sim \pi(s')}[m_{\bar \theta}(s',a',s^+)]$. As in classical RL, the gradient of the loss is not propagated through the target, and $\bar{\theta}$ refers to a stop-gradient or target network. Notably, in the following subsections we only have to deal with the Radon-Nikodym density $m^\pi$, and assume that $\rho$ is a fixed (data) distribution.

\paragraphx[-0.1em]{Relation to Framework} While the SM is not a standalone framework in itself, in its most basic formulation it can describe a reward-free, compositional method with target $\mu^\pi=m^\pi$, and the optimal value function can be recovered via the decomposition operator
\begin{equation}
    \label{eq::PolDistSM}
     \mathcal{F}(\mu^\pi,r) = \max_{\pi\in \Pi} \EE_{s^+ \sim \rho}[\mu^\pi(s,a,s^+)\, r(s^+)], %
\end{equation}
policy extraction is facilitated as $\pi_r(s) = \arg\max_a \max_{\pi\in \Pi} \EE_{s^+ \sim \rho}[\mu^\pi(s,a,s^+)\, r(s^+)]$. Notably, if $\Pi$ is sufficiently expressive, specifically if it contains the optimal policy $\pi^*$ for a given reward $r$, then full state-action coverage (support) is a sufficient condition for recovery of the optimum. In contrast, if the search space is restricted to a single SM, encoding, for example, a behavior policy $m^{\pi_b}$, coverage alone does not guarantee optimality. Assuming $m^{\pi_b}$ encodes the necessary transitions to define $\pi^*_r$, the policy extraction operation $\arg\max_a\mathbb E_{s^+\sim\rho}[m^{\pi_b}(s,a,s^+)r(s^+)]$ lacks the ability to break the temporal structure induced by the behavior policy itself, %
making optimality guarantees a function of  both temporal structure and coverage. Nonetheless, an approach following this framework, has shown good empirical performance \citep{zheng2026can}.
Lastly, it should be noted that SF and SM are related as $\psi^\pi(s,a)=\EE_{s^+ \sim \rho}[m^\pi(s,a,s^+)\phi(s^+)]$ \citep{touatidoes}.

\subsection{Forward-Backward Representation}\label{sec::FB}

Approaches that directly approximate the successor measure have the advantage of tracking successor states rather than features, thus not requiring a secondary objective to learn the basic features.
Forward-Backward representations ~\citep[FB;][]{blier2021learning,touati2021learning, touatidoes} factorize the successor measure into a low-rank decomposition:
\begin{equation}
    m^\pi(s,a,s^+) = F(s,a)^\top B(s^+),
\end{equation}
where $F: \S\times\A\to \RR^d$ and $B: \S\to\RR^d$.
As for the SM, reward-free objectives can be defined for FB, such as Eq.~(\ref{eq::SMloss}). However, naive policy-conditioned approaches may suffer from an expensive test-time search over the representation to extract the optimal policy.

Later works \citep{touati2021learning}, propose to employ a similar strategy to that of USF to FB, namely, relating policy parametrization and rewards. For a parametric family of policies $(\pi_z)_{z \in \mathcal{Z}}$, we define the forward backward representation as as a pair of functions $F: \S\times\A\times\mathcal{Z}\to\mathcal{Z}$ and $B: \S\to\mathcal{Z}$ such that for all $z\in \mathcal{Z}, (s,a) \in \S\times\A$
\begin{equation}
    \label{eq:defiFB}
    \pi_z(\cdot | s) = \arg\max_a F(s,a,z)^\top z, \quad   m^{\pi_z}(s,a,s^+) = F(s,a,z)^\top B(s^+).
\end{equation}
Crucially, for any bounded reward function $r$, by defining $z$ as the reward projection $z_r\coloneqq\EE_{s\sim \rho}[B(s)r(s)]$ we can recover the optimal value function as $Q^*_r(s,a)=F(s,a,z_r)^\top z_r$ \citep[Theorem 2 of][]{touati2021learning}. Analogously to USF, FB aligns the policy and reward spaces by conditioning a family of optimal policies on a task-specific latent $z$. However, FB departs from USF in its theoretical expressivity: whereas USF is constrained by a pre-defined set of basic features $\phi$, FB can in theory represent any reward function.

\paragraphx[-0.1em]{Training} Training is performed by minimizing a loss similar to that of Eq.~(\ref{eq::SMloss}) where we additionally sample over the space of reward embeddings $\mathcal Z$.

\begin{equation}
\label{eq::fb_loss}
    L(F,B) = 
    \EE_{s,a\sim\rho, s^+\sim\rho, z\sim p_\mathcal{Z}}[(m_{F_z,B}(s,a,s^+) - \hat m_{\bar F_z,\bar B}(s,a,s^+))^2],
\end{equation}
where $m_{F_z,B}$ is a parametric version of $m^{\pi_z}$. With $\delta_\rho(s^+) = \frac{\delta_s(\d s^+)}{\rho(\d s^+)}$, the target is calculated as
$
\hat m_{\bar F_z,\bar B}(s,a,s^+) =
\delta_\rho(s^+) +
\gamma \EE_{s'\sim p(\cdot|s,a), a'\sim \pi_z(\cdot | s')}[
\hat m_{\bar F_z,\bar B}(s',a',s^+)
]$. In continuous spaces, the hard maximum over actions is replaced by either a softmax \citep{touati2021learning} or a $z$-conditioned policy network \citep{touatidoes}. To resolve the scale ambiguity in the factorization $F^\top B$ an auxiliary orthonormalization loss enforcing $\mathbb{E}_\rho[BB^\top] \approx I$ is added \citep{touati2021learning}.

\paragraphx[-0.1em]{Relation to Framework}
In FB representations, the training target is the joint representation $\mu^r = (F,B)$. Due to the intrinsic coupling between policy and reward, training an FB representation constitutes a \emph{pseudo reward-free} approach: although it aims to generalize across rewards, it still requires access to task embeddings during training in order to optimize its objective. For a given reward function $r$, its corresponding representation can be obtained as $z_r(B) = \mathbb{E}_{s \sim \rho}[r(s) B(s)]$, which induces the following decomposition:
\begin{equation}
    \mathcal{F}(\mu=(F,B), r) = F(s,a,z_r(B))^\top z_r(B).
\end{equation}
Policy inference is then performed accordingly via $\pi_r(\cdot|s) = \arg\max_a F(s,a,z_r(B))^\top z_r(B)$.

\subsection{Proto Successor Measure}\label{sec::psm}
Proto Successor Measure \citep[PSM;][]{agarwal2025proto} takes a reward-free approach to learning the space of successor measures. Extracting the optimal policy from such representation requires a search. PSM exploits properties of the SM to define a tractable search over such space.

The SM admits a linear decomposition $m^{\pi_y}(s,a,s^+) = \Phi(s,a,s^+)^\top y + b(s,a,s^+)$, where the basis function $\Phi: \mathcal{S} \times \mathcal{A} \times \mathcal{S} \to \mathbb{R}^d$ and bias term $b: \mathcal{S} \times \mathcal{A} \times \mathcal{S} \to \mathbb{R}$ are policy-independent. The policy $\pi_y$ is represented by the weight vector $y \in \mathbb{R}^d$. This linear decomposition allows to cast the zero-shot objective under the dual linear programming  formulation (LP). Specifically, the objective is to find a $y$, that maximizes some reward function $r$. We can ensure the search is constrained to only valid successor measures by applying a non-negativity constraint, resulting in the following objective for extracting the optimal policy weights $y^*$:
\begin{equation}
\label{eq::LP_PSM}
    y^* =\arg\max_{y} \mathbb{E}_{s \sim p_0, a \sim \pi_y(\cdot|s), s^+ \sim \rho}[ m^{\pi_y}(s,a,s^+)r(s^+) ]
    \quad \text{s.t.} \quad m^{\pi_y}(s,a,s^+)   \ge 0 \quad \forall \,(s,a,s^+),
\end{equation}
The expectations over $p_0$ and $\pi_y$ are intractable since we lack access to these distributions, and are replaced by expectations over transitions from the dataset $\rho$, restricting coverage to the support of the behavior policy. The non-negativity constraint is rendered tractable by relaxing it into a soft Lagrangian penalty.

\paragraphx[-0.1em]{Training}
In discrete MDPs, the components of the representation can be learned by minimizing an objective similar to Eq.~\ref{eq::SMloss}. To allow sampling over policies, we define a discrete family of policies $(\pi_v)_{v\in \mathcal V}$, with $\mathcal V \subseteq \mathbb Z$; an additional parametric function $y: \mathbb Z \to \mathbb R^d$ maps each policy identifier onto a vector. The components are individually parametrized and learned by minimizing a reward-free objective over the dataset of transitions and policies, resulting in the following loss
\begin{equation}
\label{eq::psm_loss}
    L(\Phi,y,b) = 
    \EE_{s,a\sim\rho,\, s^+\sim\rho, v\sim p_{\mathcal V}}
    [(m_{\Phi, y(v),b}(s,a,s^+) - \hat m_{\bar \Phi, \bar y(v),\bar b}(s,a,s^+))^2],
\end{equation}
where $m_{\Phi, y(v),b}$ is a parametric version of $m^{\pi_{y(v)}}$. The target is calculated similarly to FB, Eq. \ref{eq::fb_loss}.

Given a learned representation, the policy extraction phase requires solving the LP in Eq.~\ref{eq::LP_PSM}, from which we obtain the SM of the optimal policy. %

\paragraphx[-0.1em]{Relation to Framework} The representation in PSM does not exploit ties between rewards and policies, making this a reward-free, compositional method. This also allows the representation to be learned through Bellman evaluation backups, avoiding the possible instabilities caused by the maximization over the action-value function, typical of optimality backups. This methods exhibit the same decomposition operator as the SM, defined by Eq.~\ref{eq::PolDistSM}. Based on this decomposition operator, policy extraction is performed via the search defined in Eq.~\ref{eq::LP_PSM}, resulting in the optimal parameters $y^*$ that define the extracted policy.

\section{Error Decompositions} \label{sec:error_decomp}

\begin{figure}[t]
    \centering
    \includegraphics[width=0.9\linewidth]{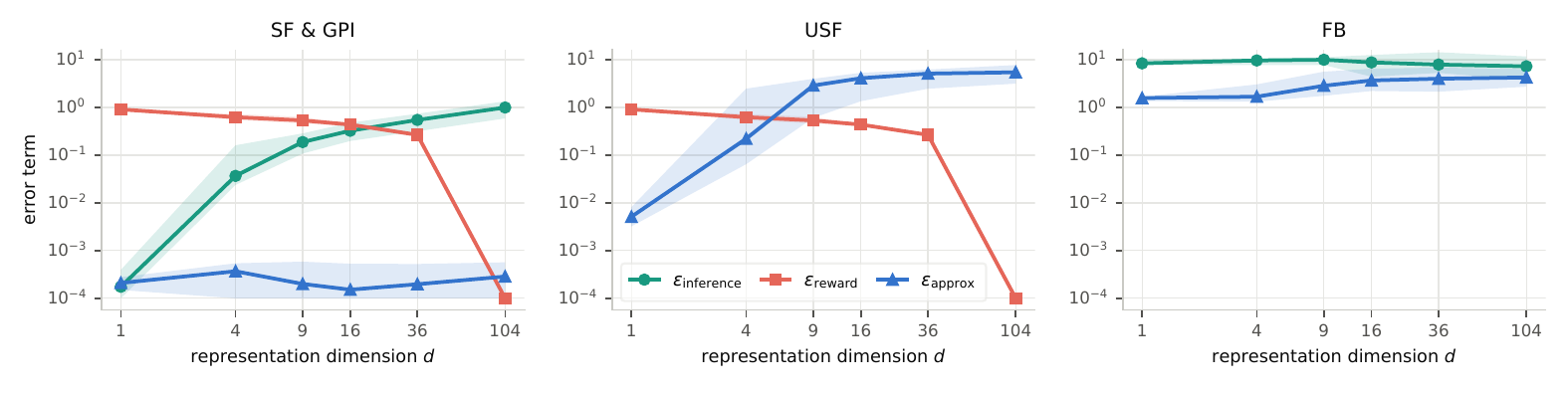}
    \caption{Error terms of Eq.~\ref{eq::error_decomposition} versus representation
    dimension $d$ in Four Rooms (log scale; median and IQR over seeds). Terms absent
    by construction are omitted. Details in Appendix~\ref{sec:appendix:empirical-error}.}
    \label{fig:empirical_bound}
\end{figure}

All previously introduced methods are theoretically well-founded and recover the optimal value function in the idealized setting via $Q_r^\ast = \mathcal{F}(\mu, r)$. While optimality is theoretically attainable, practical implementations, constrained by factors such as finite model capacity, restricted search spaces, and approximate expectations, incur three primary sources of error:
\vspace{-1em}
\begin{enumerate}
    \itemsep-0.2em 
    \item[\textcolor{custom_green}{\scalebox{1.35}{$\bullet$}}] \textbf{Inference Error:} In some cases it might not be possible to exactly evaluate the decomposition operator. For instance, when $\mathcal{F}$ requires a search over the full policy space, such as in SF\&GPI.
    \item[\textcolor{custom_red}{\scalebox{1.35}{$\bullet$}}] \textbf{Reward Error:} Even with access to the true reward function, reward embedding-based methods may suffer from errors introduced by the latent representation itself.
    \item[\textcolor{custom_blue}{\scalebox{1.35}{$\bullet$}}] \textbf{Approximation Error:} We always recover an approximate variant of the true $\mu$, due to limited data and computational resources. 
\end{enumerate}
\vspace{-1em}
In this section we build on existing results to establish error bounds for the individual zero-shot methods introduced above under a common error decomposition. Informally, given our extracted policy $\tilde{\pi} = \arg\max_a \tilde{\mathcal F}(\tilde{\mu},\tilde{r})$ we establish error bounds of the generic form
\begin{equation}
\label{eq::error_decomposition}
    \Vert Q_r^* - Q_r^{\tilde{\pi}}\Vert \leq C_1\, \underbrace{\textcolor{custom_green}{\Vert \tilde{\mathcal{F}} - \mathcal{F} \Vert}}_{\varepsilon_{\mathrm{inference}}} +\, C_2\, \underbrace{\textcolor{custom_red}{\Vert \tilde{r} - r\Vert}}_{\varepsilon_{\mathrm{reward}}} +\, C_3\, \underbrace{\textcolor{custom_blue}{\Vert \tilde{\mu} - \mu  \Vert}}_{\varepsilon_{\mathrm{approx}}}.
\end{equation}
These bounds enable a qualitative comparison of the methods presented in this work and expose the structural assumptions under which individual error terms vanish by construction. We complement this with a small-scale empirical study (cf. Figure~\ref{fig:empirical_bound}), which confirms that this behavior is observable in practice, but also that there is no free lunch: methods incur comparable total error and merely trade off its sources. 
For readability, the main text states reduced versions of the theorems; full statements, proofs, exact constants, and norm definitions are deferred to Appendix~\ref{sec:appendix:proofs}.

\textbf{Successor Features \& GPI\,} We extend the results of \citep{barreto_successor_2017, borsa2018universal} to account for approximate reward linearizations:

\begin{theorem}
    \label{thm:1}
    Let $\phi$ be a bounded basic feature mapping, 
    let $(\mathcal{W}, \cdot^\top\cdot)$ be an inner product space, 
    let $\{w_i\}_{i=1}^n \in \mathcal{W}$ be linear weights and let $\{\pi_i\}_{i=1}^n$ be a set of the corresponding policies, such that $\pi_i$ is optimal on $\phi^\top w_i$. Let $\tilde{\psi}_i(s,a)$ denote the learned SF for the given policies. Let $r \in \mathcal{R}$ be a bounded reward function and $w_r$ its approximate linearization such that $r \approx \phi^\top w_r$, and let $\pi(\cdot|s) = \arg\max_a \max_{i \in [n]} \tilde{\psi}_i(s,a)^\top w_r$ be the GPI policy. Then there exists positive constants $C_{\gamma,\phi}, C_\gamma, C_r $ such that 
    \begin{align}
        \label{eq::thm1}
        \Vert Q^*_r  - Q^\pi_r\Vert_\infty \leq C_{\gamma,\phi} \,\textcolor{custom_green}{\min_{i \in [n]} \Vert w_r - w_i\Vert} +  C_\gamma\, \textcolor{custom_red}{\Vert r - \phi^\top w_r\Vert} + C_{r}\, \textcolor{custom_blue}{\max_{i \in [n]} \Vert \psi^{\pi_i}(\cdot) - \tilde{\psi}_i(s,a)\Vert}.
    \end{align}
\end{theorem}
The first term reflects a restricted policy search: unless an optimal policy for the target reward is contained in the training set, an error during policy extraction is unavoidable. This error can be expected to decrease as the number of trained policies grows. The second term captures reward linearization error, which vanishes only if rewards are exactly linear in $\phi$. The final term corresponds to approximation error in learning successor features, as a result from finite model capacity and limited training data.

\textbf{Universal Successor Features\,} In contrast to SF\&GPI, USFs remove explicit inference error by parameterizing policies through reward weights. However, generalization across $\mathcal W$ may increase approximation error, especially in regions where coverage is limited. As a consequence of Theorem \ref{thm:1} \citep[using the argument in][]{borsa2018universal} we obtain the following error bound for USFs:

\begin{theorem}
    \label{thm:2}
    Let $\phi$ be a bounded basic feature mapping, 
    let $(\mathcal{W}, \cdot^\top\cdot)$ be an inner product space, 
    let $\tilde{\psi}(\cdot, w)$ denote the learned USFs for some $w \in \mathcal{W}$.  Let $r \in \mathcal{R}$ be a bounded reward function and $w_r \in \mathcal{W}$ its approximate linearization such that $r \approx \phi^\top w_r$, and let $\pi(\cdot|s) = \arg\max_a \tilde{\psi}(s,a,w_r)^\top w_r$ be the extracted policy. Let $\psi^{w_r}(\cdot)$ denote the SFs for the policy optimal on $\phi^\top w_r$. Then:
    \begin{align}
        \label{eq::thm2}
        \Vert Q^*_r  - Q^\pi_r\Vert_\infty \leq  C_\gamma\, \textcolor{custom_red}{\Vert r - \phi^\top w_r\Vert} + C_{r}\, \textcolor{custom_blue}{\Vert \psi^{w_r}(\cdot) - \tilde{\psi}(\cdot, w_r)\Vert}.
    \end{align}
\end{theorem}

\textbf{Forward-Backward Representation\,} A key distinction of FB representations is that they parametrize the successor measure without imposing any assumption on the reward function, which is reflected in their error bound. Extending the results of \citet{touati2021learning} by an approximation error term, we obtain:

\begin{theorem}
    \label{thm:3}
    Let $\Pi_\mathcal{M}$ denote the set of all memory-less policies on our family of MDPS $\mathcal{M}^{\mathcal{R}}$. For any $\pi \in \Pi_\mathcal{M}$ let $m^\pi$ denote the successor measure. Let $\mathcal{Z}$ be an inner product space and let $F,B$ a forward-backward representation, that is a pair of functions $F: \S\times\A\times\mathcal{Z}\to\mathcal{Z}$ and $B: \S\to\mathcal{Z}$, with factorization defect $\varepsilon_{\pi,z}(s,a,s^+) = m^{\pi}(s,a,s^+) - F(s,a,z)^\top B(s^+)$. Let $\tilde F, \tilde B$ denote their learned versions. Let $r \in \R$ be a bounded reward function, let $\rho \in \Delta(\S)$ a distribution of states, let $z_r = \EE_{s \sim \rho}[B(s)r(s)]$ and let $\pi(\cdot|s) = \arg\max_a \tilde F(s,a,z_r)^\top z_r$, then there exists positive constants $C_{r,\gamma},\tilde{C}_{r,\gamma}$ only dependent on $r$ and $\gamma$ such that 
    \begin{equation}
        \label{eq:thm3}
        \Vert Q^*_r  - Q^\pi_r\Vert_\infty \leq 
        C_{r,\gamma}\, \textcolor{custom_green}{\left(\Vert \varepsilon_{\pi_{z_r},z_r}\Vert + \Vert \varepsilon_{\pi,z_r}\Vert\right)}
        + \tilde{C}_{r,\gamma}\,\textcolor{custom_blue}{\left(\Vert F(\cdot,z_r) - \tilde{F}(\cdot,z_r)\Vert + \Vert F^\top B - \tilde F^\top \tilde B \Vert \right)}.
    \end{equation}
\end{theorem}

The theorem leaves the reference point $F, B$ unspecified. We regard it as the ground truth, i.e. the factorization with minimal (ideally vanishing) defect. Under this reading, the first term in Eq.~\ref{eq:thm3} captures the inference error induced by the factorization defect, that is, the error incurred by not representing the successor measure exactly under the given parametrization. While this inference error is conceptually distinct from the linearization error in USFs, the two are closely related \citep[see][for a detailed discussion]{touati2021learning}. 
Unlike USFs, FB does not impose a linear decomposition of the reward function apriori, and therefore incurs no explicit reward approximation error of the form appearing in Eqs.~\ref{eq::thm1} and \ref{eq::thm2}.

We conjecture that PSM  admits a similarly-composed error bound to FB, where the inference would also be tied to the soft LP. %

\begin{example}
    To illustrate these bounds in practice, we instantiate the decomposition of Eq.~\ref{eq::error_decomposition} in a deterministic Four Rooms grid-world ($|\S| = 104$, $|\A| = 4$, $\gamma = 0.95$), small enough that every quantity in Theorems~\ref{thm:1}--\ref{thm:3} is computed rather than estimated, so that each reported error is attributable to the representation alone. SF-based methods use one-hot block indicators as features $\phi$; USF and FB additionally require a model of how $\psi$ varies with the task, for which we use a bilinear model with one RBF per training task;
    see Appendix~\ref{sec:appendix:empirical-error} for details.

    Figure~\ref{fig:empirical_bound} shows the resulting error curves over increasing representation dimension $d$. There is indeed no free lunch: methods trade error between sources while total error remains of comparable magnitude. USF and SF\&GPI incur the same reward error, both assuming a linear reward representation; since $\phi$ is a block indicator, larger $d$ yields finer blocks and eventually enough capacity to represent every reward function linearly, so this error vanishes. FB incurs no reward error at all. The approximation error is markedly larger for USF and FB than for SF\&GPI, which fits a separate object per policy rather than generalizing across the representation space, and it grows with $d$ for the latter two. Finally, the inference error dominates for FB and, at larger $d$, for SF\&GPI, whereas USF incurs none by construction. For FB it decreases with $d$, as added capacity reduces the factorization defect; for SF\&GPI with a fixed number of policies $n$, it increases with $d$, since the same policy set covers an increasingly large representation space. These terms also localize where a distribution shift in the reward functions takes effect. For USF and FB it enters through the approximation error: both generalize across tasks, so their fit degrades on rewards far from those seen during training. SF\&GPI is trained reward-free and is therefore not affected directly; the shift is instead captured in the inference error, as a mismatch between the rewards whose optimal policies are contained in the baseline set and those encountered at test time.

\end{example}

\textbf{Direct Methods\,} Direct approaches do not rely on value-function decomposition and therefore incur no decomposition-induced inference error:

\begin{theorem}
    \label{thm:5}
Let $\Z\subset \RR^d$ and let $f: \R \to \Z\subset \RR^d$ be a reward embedding. For $z \in \Z$ let $\tilde{Q}_z$ denote some trained direct representation. Let $r \in \R$  be a bounded reward function and $f(r)=z_r$ its reward embedding. Then the value error of direct methods decomposes into
\begin{equation}
    \Vert Q_r^\ast - \tilde{Q}_{f(r)}\Vert_\infty \leq \inf_{\tilde{r} \in f^{-1}(\{z_r\})} \left(\frac{1}{1 - \gamma} \textcolor{custom_red}{\Vert r - \tilde{r} \Vert} + \textcolor{custom_blue}{\Vert Q^*_{\tilde{r}} - \tilde{Q}_{z_r}\Vert} \right).
\end{equation}
\end{theorem}

In general, the pre-image of a reward embedding contains more than one element:
compressing the reward space into a latent space $\Z$ forces distinct reward
functions onto the same embedding. A single $\tilde{Q}_z$ can then not accurately
approximate all $Q^*_r$ with $r \in f^{-1}(\{z\})$ simultaneously, and the severity
of this effect is affected by how much these optimal value functions differ across
the pre-image. Non-identifiability thus comes at a price: even if every reward
function is representable in the latent space (which is not guaranteed, so a
reward error may already be incurred) and the reward error can be driven to zero,
the approximation error is expected to grow with the degree of compression of $f$.

\section{Related Work}

Most closely related to our work is the unified framework of \cite{agarwal2025unified}, which establishes a formal equivalence of several unsupervised RL objectives via the SM. We instead structure the landscape of zero-shot methods by analyzing the implications of various design choices, offering a complementary view. %

\emph{Unsupervised RL} methods \citep{jaderberg2017reinforcement,eysenbach2019diversity,laskin2021urlb} undergo a pretraining phase (either in an reward-free or pseudo reward-free paradigm) followed by a finetuning phase to an extrinsic reward. Zero-shot RL is a subset of URL, where the finetuning phase is ignored and the agent is evaluated immediately after the pretraining phase.

\emph{Transfer learning} in RL (TL), as surveyed in \cite{taylor2009transfer}, improves sample efficiency and generalization across tasks by reusing knowledge from prior experience. TL encompasses paradigms that adapt 
to changes in all parts of the MDP. Within this broader framework, zero-shot RL can be viewed as a form of transfer where adaptation is restricted to changes in the reward. 

\emph{Model-based RL}, as surveyed in~\cite{moerland2023model}, leverages transition dynamics models to plan by unrolling simulated trajectories \citep{janner2019trust}. While decoupling dynamics from rewards enables flexibility, determining optimal values requires tracing potential futures. Model-based agents therefore adapt to new rewards through test-time search, while zero-shot RL demands fast generalization to novel objectives without explicit test-time planning.

\section{Discussion \& Conclusion}

We introduced a unified framework for zero-shot RL, providing structure to a previously fragmented field. Our taxonomy distinguishes methods by representation structure, direct versus compositional, and by learning paradigm, separating reward-free from pseudo reward-free objectives. A unified error decomposition further separates inference, reward, and representation approximation error.

This work also highlights several directions for future research. First, direct methods may benefit from advances in representation learning, where smoother reward embeddings could improve generalization \citep{bengio2013representation,wu2018unsupervised}. In continuous spaces, greedy policy extraction may introduce out-of-distribution errors, which regularization may mitigate \citep{jeen2024zero,zheng2025towards}. Reward-free compositional methods instead rely on Bellman evaluation backups, offering a possible way to avoid such compounding errors. While most recent works focus on offline RL to isolate the representation learning problem, the online regime introduces exploration challenges; leveraging zero-shot representations for exploration remains a promising direction \citep{sun2025unsupervised, urpi2025epistemically}. Finally, existing benchmarks such as URLB \citep{laskin2021urlb} or ExoRL \citep{yarats2022exorl} may obscure representation-specific limitations, motivating the development of dedicated benchmarks \citep{cetin2025finer}.

\bibliography{main}
\bibliographystyle{tmlr}
\newpage
\appendix

\section{Feature Matrix of Zero-Shot Methods}
\label{sec::app-featurematrix0srlMethods}
\begin{table}[H]
\centering
\begin{threeparttable}
\label{table:feature_matrix_table}
\small %
\renewcommand{\arraystretch}{1.2}
\rowcolors{2}{gray!5}{white} %
\caption{Feature matrix of a \emph{non-exhaustive} list of zero-shot RL methods}
\begin{tabular}{
@{} 
p{2.2cm}
c 
c 
c 
c 
@{}
}
\toprule
\textbf{Method} & 
\textbf{Representation} & 
\textbf{Reward} & 
\textbf{Paradigm} & 
\makecell{\bf Policy\\ \bf Extraction} \\ 
\midrule
SF+GPI \tiny{\citep{barreto_successor_2017}} & Compositional & Free & Online & GPI\\
USF \tiny{\citep{borsa2018universal}} & Compositional & Pseudo & Online & Instant \\
USF Lap. \tiny{\citep{wu2019laplacian}} & Compositional & Pseudo & Offline & Instant \\
FB \tiny{\citep{touati2021learning,touatidoes}} & Compositional & Pseudo & Offline & Instant \\
FB-AWARE \tiny{\citep{cetin2025finer}} & Compositional & Pseudo & Offline & Instant \\
VISR \tiny{\citep{hansen_fast_2020}} & Compositional & Pseudo & Online & Instant \\
APS \tiny{\citep{liu2021aps}} & Compositional & Pseudo & Online & Instant \\
PSM \tiny{\citep{agarwal2025proto}} & Compositional & Free & Offline & Constrained LP \\
FRE \tiny{\citep{frans2024unsupervised}} & Direct & Pseudo & Offline & Instant \\
HILP \tiny{\citep{park2024foundation}} & Direct & Pseudo & Offline & Instant \\
FBEE$^Q$ \tiny{\citep{urpi2025epistemically}} & Compositional & Pseudo & Online & Instant \\
DVFB \tiny{\citep{sun2025unsupervised}} & Compositional & Pseudo & Online & Instant \\
VC-FB \tiny{\citep{jeen2024zero}} & Compositional & Pseudo & Offline & Instant \\
BREEZE \tiny{\citep{zheng2025towards}} & Compositional & Pseudo & Offline & Instant \\
TD-JEPA \tiny{\cite{bagatella2026tdjepa}} & Compositional & Pseudo & Offline & Instant \\
one-step FB \tiny{\citep{zheng2026can}} & Compositional & Free & Offline & Instant \\
\bottomrule
\end{tabular}
\end{threeparttable}
\end{table}

\begin{remark}
    According to our framework, any reward-free method that extracts a policy with
    optimality guarantees (under the usual assumptions) for a given reward must
    search over policies using the learned representation $\mu^\pi$; instant
    extraction is possible only when policies are already parametrized by the reward
    for which they are optimal, which occurs only in pseudo reward-free settings.
    One-step FB~\citep{zheng2026can} may appear to be a counterexample: it is
    reward-free yet extracts a policy instantly. It does so by learning a single
    successor measure $m^{\pi_b}$ for the behavior policy $\pi_b$ and defining
    \[
        \pi_r(s) = \argmax_{a}\; \EE_{s^+ \sim \rho}\!\left[\, m^{\pi_b}(s, a,s^+)\, r(s^+) \,\right].
    \]
    This does not contradict our taxonomy: the resulting policy $\pi_r$ is obtained via a single step of policy improvement over $m^{\pi_b}=\mu^{\pi_b}$, as such it carries no guarantees of optimality for the reward $r$. Nonetheless, its categorization under the proposed two-axes taxonomy remains correct: the representation $\mu^{\pi_b}$, is indeed learned in a reward free setting, as it simply encodes the behavior policy inferred from the dataset. Instant extraction is achieved by one-step improving $\mu^{\pi_b}$ for each reward in $\mathcal D^{\text{train}}$, yielding a task-conditioned policy $\pi(\cdot|s,r)$.
\end{remark}
\begin{remark}
    This PSM categorization applies only for its discrete action-space instantiation. PSM is a reward-free method only in the discrete action space setting, in the continuous setting the method becomes a feature learning algorithm for USF \citep{agarwal2025proto}.
\end{remark}

\newpage

\section{Notation}
\label{sec::app-Notation}

We generally assume that any reward function $r \in \R$ is a mapping $r: \S \times \A  \times \S \to \mathbb{R}, (s,a,s') \mapsto r(s,a,s')$. In some MDPs, the dependence on  $a$ or $s'$ entirely drops, as for instance in goal-conditioned RL. When full dependence is present, we use the following short-hand notations by marginalizing over action and future state distributions
\begin{align*}
    r(s,a) &=\mathbb{E}_{s' \sim p(\cdot|s,a)}[r(s,a,s')],\\
    r(s) &=\mathbb{E}_{a\sim \pi(\cdot|s), s' \sim p(\cdot|s,a)}[r(s,a,s')] .
\end{align*}

\section{Proofs}\label{sec:appendix:proofs}

Throughout this section we fix an arbitrary family of MDPs $\mathcal{M}^{\mathcal{R}}$,
where $\mathcal{R}$ denotes a set of bounded reward functions; all statements below are
to be understood with respect to this family.
For a bounded function $f \colon \X \to (N, \lvert \cdot \rvert)$ taking values in a
normed space, we write
\begin{equation*}
  \lVert f \rVert_\infty \coloneqq \sup_{x \in \X} \lvert f(x) \rvert ,
\end{equation*}
which defines a norm on the vector space $B(\X, N)$ of bounded functions from $\X$ to $N$.

\subsection{Theorem \ref{thm:1}}

\begin{theorem}[Theorem \ref{thm:1}]
    Let $\phi$ be a bounded basic feature mapping, let $(\mathcal{W}, \cdot^\top\cdot)$ be an inner product space, let $\{w_i\}_{i=1}^n \in \mathcal{W}$ be linear weights and let $\{\pi_i\}_{i=1}^n$ be a set of the corresponding policies, such that $\pi_i$ is optimal on $\phi^\top w_i$. Let $\tilde{\psi}_i(s,a)$ denote the learned SF for the given policies. Let $r \in \mathcal{R}$ be a bounded reward function and $w_r$ its approximate linearization such that $r \approx \phi^\top w_r$, and let $\pi(\cdot|s) = \arg\max_a \max_{i \in [n]} \tilde{\psi}_i(s,a)^\top w_r$ be the GPI policy. Then 
    \begin{align*}
        \Vert Q^*_r  - Q^\pi_r\Vert_\infty \leq \frac{2}{1-\gamma}\min_{i \in [n]}\Vert \phi\Vert_\infty \Vert w_r - w_i\Vert +  \frac{2}{1-\gamma} \Vert r - \phi^\top w_r\Vert_\infty + \max_{i \in [n]} \Vert w_r \Vert \Vert \psi^{\pi_i}(\cdot) - \tilde{\psi}_i(s,a)\Vert_\infty,
    \end{align*}
    where $\Vert \cdot \Vert$ is the norm induced by the inner product on $\mathcal{W}$.
\end{theorem}

\begin{proof}
    We can decompose the optimality gap as
    \begin{align}
        \notag
        \Vert Q^*_r  - Q^\pi_r\Vert_\infty 
        &\leq \Vert Q^*_r  - Q^*_{\phi^\top w_r}\Vert_\infty + \Vert Q^*_{\phi^\top w_r}  - Q^{\pi}_{\phi^\top w_r}\Vert_\infty + \Vert Q^{\pi}_{\phi^\top w_r}  - Q^\pi_r\Vert_\infty \\
        \label{eq:proof1-1}
        & \leq 2 \max_{\tilde{\pi}} \Vert Q^{\tilde{\pi}}_{\phi^\top w_r}  - Q^{\tilde{\pi}}_r\Vert_\infty + \Vert Q^*_{\phi^\top w_r}  - Q^{\pi}_{\phi^\top w_r}\Vert_\infty.
    \end{align}
    For clarity of notation we write $\sup_{(s,a) \in \S\times\A}|Q(s,a) = \Vert Q\Vert_\infty$. Note that $Q$ is bounded as we assume a bounded reward function.

    \noindent Here the first inequality follows from the triangle inequality. Lastly, by Theorem~1 in~\cite{borsa2018universal} we have
    \begin{equation}
        \label{eq:proof1-2}
        \Vert Q^*_{\phi^\top w_r}  - Q^{\pi}_{\phi^\top w_r}\Vert_\infty \leq \frac{2}{1-\gamma}\min_{i \in [n]}\Vert \phi\Vert_\infty \Vert w_r - w_i\Vert + \max_{i \in [n]} \Vert w_r \Vert \Vert \psi^{\pi_i} - \tilde{\psi}_i\Vert_\infty,
    \end{equation}
    where $\Vert \cdot \Vert$ is the norm induced by the inner product on $\mathcal{W}$, that is $\Vert w \Vert = (w^\top\,w)^{1/2}$, and $\Vert \phi\Vert_\infty = \sup_{(s,a) \in \S\times\A}||\phi(s,a)||$. Notably, a bounded $\phi$ implies the boundedness of $\psi$. 

    For any policy $\pi$ and two reward functions $r,r'$ we can estimate the gap of the value functions using the result of Proposition 17 from \citep{touati2021learning}:
    \begin{equation}
        \label{eq:prop17:touati}
        \Vert Q_r^\pi - Q_{r'}^\pi\Vert_\infty \leq \frac{1}{1-\gamma} \Vert r - r' \Vert_\infty,
    \end{equation}
    where similarly to before we write $\Vert r\Vert_\infty = \sup_{(s,a) \in \S\times\A}|r(s,a)$.
    This, in combination with Eq. (\ref{eq:proof1-2}) and (\ref{eq:proof1-1}) completes the proof.

\end{proof}

Note, that in finite MDPs, we can also use the euclidean norm to bound this value-gap, instead of using the supremum: For any policy $\pi$, and reward functions $r,r'$ we have
\begin{equation}
    \label{eq::LinearizationErrorBound}
    \Vert Q_r^\pi - Q_{r'}^\pi \Vert_\infty \leq \frac{\Vert r-r'\Vert_2}{1-\gamma}.
\end{equation}
Here $r,r'$ are treated as vectors in $\RR^{|\S|\times|\A|}$, and $\Vert\cdot\Vert_2$ denotes the euclidean norm. Notably, for $r' = \phi^\top w_r$ this distance is indeed minimized by obtaining $w_r$ that minimizes the mean squared error, which can be obtained by sampling uniformly in Eq. (\ref{eq::lin_reg_task_sf}).

\begin{proof}
    For convenience of notation we write $\Delta r(\cdot) = r(\cdot) - r'(\cdot)$. Moreover, let $\rho^\pi(\cdot|t,s,a) \in \Delta(\S\times\A)$ denote the state-action distribution under $\pi$ at timestep $t$ starting in $s_0=s,a_0=a$. As we are in the finite MDP setting we will treat $\Delta r$ and $\rho^\pi(\cdot|t,s,a)$ as objects in $\RR^{|\S|\times|\A|}$, and denote by $\langle \cdot,\cdot \rangle$ the standard inner product on this space.
    
    For some policy $\pi$ and an arbitrary state action pair $(s,a) \in \S\times\A$ we have
    \begin{align*}
        &|Q_r^\pi(s,a) - Q_{r'}^\pi(s,a)| \\
        &\leq \sum_{t \geq 0} \gamma^t |\langle \rho^\pi(\cdot|t,s,a), \Delta r(\cdot)\rangle | \\
        &\leq \sum_{t \geq 0} \gamma^t \Vert\rho^\pi(\cdot|t,s,a)\Vert_2 \Vert \Delta r\Vert_2 \\
        &\leq \Vert \Delta r\Vert_2 \sum_{t \geq 0} \gamma^t \Vert\rho^\pi(\cdot|t,s,a)\Vert_2 \\
        &\leq \Vert \Delta r\Vert_2 \frac{1}{1-\gamma}.
    \end{align*}

    \noindent Here the second inequality follows from Cauchy-Schwarz inequality, and the final one from Cauchy-Schwarz and the fact that the distribution has bounded mass:
    \begin{equation*}
       \Vert\rho^\pi(\cdot|t,s,a)\Vert_2 \leq \Vert 1 \Vert_\infty \Vert\rho^\pi(\cdot|t,s,a)\Vert_1 \leq 1.
    \end{equation*}

    The established expression is independent of the chosen state-action pair completing the proof.
\end{proof}

\subsection{Theorem \ref{thm:2}}

\begin{theorem}[Theorem \ref{thm:2}]
    Let $\phi$ be a bounded basic feature mapping, let $(\mathcal{W}, \cdot^\top\cdot)$ be an inner product space, let $\tilde{\psi}(\cdot, w)$ denote the learned USFs for some $w \in \mathcal{W}$.  Let $r \in \mathcal{R}$ be a bounded reward function and $w_r \in \mathcal{W}$ its approximate linearization such that $r \approx \phi^\top w_r$, and let $\pi(\cdot|s) = \arg\max_a \tilde{\psi}(s,a,w_r)^\top w_r$ be the extracted policy. Let $\psi^{w_r}(\cdot)$ denote the SFs for the policy optimal on $\phi^\top w_r$. Then:
    \begin{align}
        \Vert Q^*_r  - Q^\pi_r\Vert_\infty \leq  \frac{2}{1-\gamma} \Vert r - \phi^\top w_r\Vert_\infty  + \Vert w_r \Vert \Vert \psi^{\pi_i} - \tilde{\psi}(\cdot,w_r)\Vert_\infty,
    \end{align}
    where $\Vert \cdot \Vert$ is the norm induced by the inner product on $\mathcal{W}$.
\end{theorem}

We provide a proof for completeness; it follows the argument of
\citet{borsa2018universal}.

\begin{proof}
    The setup for USFs change in the following way: Instead of finite set $\{w_i\}$ of linear weights, we assume that we can represent the full linear weight space $\mathcal W$. We then denote $\pi^w$ as the policy that is optimal w.r.t. the rewards $\phi^\top\,w$, and let $\psi^w\coloneqq\psi^{\pi_w}$ denote the corresponding SFs, and $\tilde{\psi}(\cdot,w)$ its approximation. 

    The difference compared to SFs is now, that for any given $w_r \in \mathcal{W}$ we can immediately query $\tilde{\psi}(s,a,w_r)$, as opposed to approximate the otpimal values w.r.t $w_r$ by some finite set of linear weights, as done with SFs in Theorem \ref{eq::thm1}: $\max_{i \in [n]}\tilde{\psi}_i(s,a)^\top\,w_r$. In other words, if we extend the SF \& GPI notation such that $\tilde{\psi}_i(s,a) = \tilde{\psi}(s,a,w_i)$, the policy in SF \& GPI is extracted as 
    \begin{equation}
        \pi_{SF\&GPI}(\cdot|s) = \arg\max_a \max_{w \in \mathcal{W}_n}\tilde{\psi}(s,a,w)^\top\,w_r,
    \end{equation}
    where $\mathcal{W}_n = \{w_1,\dots,w_n\}$ is a pre-defined set of weights, that is independent of the linear weights $w_r$ exposed to during inference. In USF the situation changes, in that the min is taken over a set of weights that (i) depends on $w_r$ and (ii) contains a single element, i.e. 
    \begin{equation}
        \pi_{USF}(\cdot|s) = \arg\max_a \tilde{\psi}(s,a,w_r)^\top\,w_r = \arg\max_a \max_{w \in \{w_r\}}\tilde{\psi}(s,a,w)^\top\,w_r.
    \end{equation}

    We can hence apply the error bound of Theorem \ref{thm:1} for USF as
    \begin{align}
        \Vert Q^*_r  - Q^\pi_r\Vert_\infty &\leq \frac{2}{1-\gamma} \Vert \phi \Vert_\infty \,\min_{w \in \{w_r\}} \Vert w_r - w_i\Vert +  \frac{2}{1-\gamma}\, \Vert r - \phi^\top w_r\Vert_\infty + \Vert w_r \Vert\, \max_{w \in \{w_r\}} \Vert \psi^{w} - \tilde{\psi}(\cdot,w)\Vert_\infty \\
        &= \frac{2}{1-\gamma}\, \Vert r - \phi^\top w_r\Vert_\infty + \Vert w_r \Vert\, \Vert \psi^{w_r} - \tilde{\psi}(\cdot,w_r)\Vert_\infty
    \end{align}
\end{proof}

\subsection{Theorem \ref{thm:3}}

\begin{theorem}[Theorem \ref{thm:3}]
    Let $\Pi_\mathcal{M}$ denote the set of all memory-less policies on our family of MDPS $\mathcal{M}^{\mathcal{R}}$. For any $\pi \in \Pi_\mathcal{M}$ let $m^\pi$ denote the successor measure. Let $\mathcal{Z}$ be an inner product space and let $F,B$ a forward-backward representation, that is a pair of functions $F: \S\times\A\times\mathcal{Z}\to\mathcal{Z}$ and $B: \S\to\mathcal{Z}$, with factorization defect $\varepsilon_{\pi,z}(s,a,s^+) = m^{\pi}(s,a,s^+) - F(s,a,z)^\top B(s^+)$. Let $\tilde F, \tilde B$ denote their learned versions. Let $r \in \R$ be a bounded reward function,let $z_r = \EE_{s \sim \rho}[B(s)r(s)]$ and let $\pi(\cdot|s) = \arg\max_a \tilde F(s,a,z_r)^\top z_r$, then there exists positive constants $C_{r,\gamma},\tilde{C}_{r,\gamma}$ only dependent on $r$ and $\gamma$ such that 
    \begin{equation}
        \begin{aligned}
        &\Vert Q^*_r  - Q^\pi_r\Vert_\infty \\
        &\leq C_{r,\gamma} \left(\Vert \varepsilon_{\pi_{z_r},z_r}\Vert_\infty +  \Vert\varepsilon_{\pi,z_r}\Vert_\infty\right)+ \tilde{C}_{r,\gamma}\, \Vert F(\cdot,z_r) - \tilde{F}(\cdot,z_r)\Vert_\infty + \Vert F(\cdot,z_r)^\top B - \tilde F(\cdot,z_r)^\top \tilde B\Vert_\infty.
        \end{aligned}
    \end{equation}
    Here we denote
    \begin{align}
        \Vert F(\cdot,z_r) - \tilde{F}(\cdot,z_r)\Vert_\infty &= \sup_{s,a}\Vert F(s,a,z_r) - \tilde{F}(s,a,z_r)\Vert  \\
        \Vert \varepsilon_{\pi,z}\Vert_\infty&= \sup_{s,a}\Vert\varepsilon_{\pi,z}(s,a,\cdot)\Vert_{L^1(\rho)}  \\
         \Vert F(\cdot,z_r)^\top B - \tilde F(\cdot,z_r)^\top \tilde B\Vert_\infty&= \,\sup_{s,a} \Vert F(s,a,z_r)^\top B(\cdot) - \tilde F(s,a,z_r)^\top \tilde B(\cdot)\Vert_{L^1(\rho)},
    \end{align}
    and $\Vert \cdot \Vert$ denotes the induced norm by the inner product on $\mathcal{Z}$.
\end{theorem}

\begin{proof}
    We can firstly decompose the error as 
    \begin{equation}
        \Vert Q^*_r  - Q^\pi_r\Vert_\infty \leq \Vert Q^*_r  - Q^{\pi_{z_r}}_r\Vert_\infty + \Vert Q^{\pi_{z_r}}_r  - Q^\pi_r\Vert_\infty.
    \end{equation}

    Following \cite{touati2021learning}, we define $\varepsilon_{\pi,z}(s,a,s^+) = m^{\pi}(s,a,s^+) - F(s,a,z)^\top B(s^+)$ and $\tilde \varepsilon_{\pi,z}(s,a,s^+) = m^{\pi}(s,a,s^+) - \tilde F(s,a,z)^\top \tilde B(s^+)$. Using these definitions, the first term can be bounded in terms of the factorization defect as
    \begin{equation}
        \Vert Q^*_r  - Q^{\pi_{z_r}}_r\Vert_\infty \leq \frac{3 \Vert r\Vert_\infty}{1-\gamma} \sup_{s,a} \Vert \varepsilon_{\pi_{z_r},z_r}(s,a,\cdot)\Vert_{L^1(\rho)} =: \frac{3 \Vert r\Vert_\infty}{1-\gamma} \Vert\varepsilon_{\pi_{z_r},z_r} \Vert_\infty.
    \end{equation}

    Here $\Vert \cdot \Vert_{L^1(\rho)}$ denotes the L1-Norm. The second term can now be further decomposed as
    \begin{align*}
        \Vert Q^{\pi_{z_r}}_r  - Q^\pi_r\Vert_\infty \leq \Vert Q^{\pi_{z_r}}_r - F(\cdot,z_r)^\top z_r\Vert_\infty + \Vert Q^{\pi}_r - \tilde{F}(\cdot,z_r)^\top z_r\Vert_\infty + \Vert F(\cdot,z_r)^\top z_r - \tilde{F}(\cdot,z_r)^\top z_r\Vert_\infty.
    \end{align*}

    The first component on the right hand side can again be bounded by results from \citep{touati2021learning}:
    \begin{equation}
        \Vert Q^{\pi_{z_r}}_r - F(\cdot,z_r)^\top z_r\Vert_\infty \leq \Vert r \Vert_\infty \sup_{s,a}\Vert \varepsilon_{\pi_{z_r},z_r}(s,a,\cdot)\Vert_{L^1(\rho)} = \Vert r \Vert_\infty \Vert \varepsilon_{\pi_{z_r},z_r} \Vert_\infty.
    \end{equation}
    The same result can be applied to bound the second term as
    \begin{equation}
        \Vert Q^{\pi}_r - \tilde{F}(\cdot,z_r)^\top z_r\Vert_\infty \leq \Vert r \Vert_\infty \sup_{s,a}\Vert \tilde \varepsilon_{\pi,z_r}(s,a,\cdot)\Vert_{L^1(\rho)} = \Vert r \Vert_\infty \Vert \tilde \varepsilon_{\pi,z_r} \Vert_\infty.
    \end{equation}

    Lastly, the last component can be bounded by Cauchy-Schwartz as 
    \begin{align}
        \Vert F(\cdot,z_r)^\top z_r - \tilde{F}(\cdot,z_r)^\top z_r\Vert_\infty &= \sup_{s,a}\vert F(s,a,z_r)^\top z_r - \tilde{F}(s,a,z_r)^\top z_r\vert \\
        &\leq \Vert z_r\Vert\, \sup_{s,a}\Vert F(s,a,z_r) - \tilde{F}(s,a,z_r)\Vert =: \Vert z_r\Vert \Vert F(\cdot,z_r)\Vert_\infty,
    \end{align}
    where $\Vert \cdot \Vert$ denotes the norm induced by the inner product. Combining these yields
    \begin{align}
        \label{eq:40}
        & \Vert Q^*_r  - Q^\pi_r\Vert_\infty \leq C_{r,\gamma} \left(\Vert \varepsilon_{\pi_{z_r},z_r}\Vert_\infty+  \Vert \tilde \varepsilon_{\pi,z_r}\Vert_\infty\right) + C_r\, \Vert F(\cdot,z_r) - \tilde{F}(\cdot,z_r)\Vert_\infty,
    \end{align}
    where $C_{r,\gamma} \coloneqq \left(\frac{3}{1-\gamma} + 1 \right)  \Vert r\Vert_\infty$, and $C_r \coloneqq \Vert z_r\Vert$. Lastly, we decompose $\Vert \varepsilon_\pi(s,a,\cdot)\Vert_{L^1(\rho)}$ into terms that we can attribute to the inference and approximation error respectively. By the triangle inequality we have
    \begin{align}
        &\Vert \tilde{\varepsilon}_{\pi,z_r}(s,a,\cdot)\Vert_{L^1(\rho)} \\
        &\leq \Vert m^\pi(s,a,\cdot) - F(s,a,z_r)^\top B(\cdot)\Vert_{L^1(\rho)} + \Vert F(s,a,z_r)^\top B(\cdot) - \tilde F(s,a,z_r)^\top \tilde B(\cdot)\Vert_{L^1(\rho)}\\
        &= \Vert \varepsilon_{\pi,z_r}(s,a,\cdot)\Vert_{L^1(\rho)} + \Vert F(s,a,z_r)^\top B(\cdot) - \tilde F(s,a,z_r)^\top \tilde B(\cdot)\Vert_{L^1(\rho)}.
    \end{align}

    Combining this with Eq. \ref{eq:40} completes the proof.
    
\end{proof}

\subsection{Theorem \ref{thm:5}}

\begin{theorem}[Theorem \ref{thm:5}]
Let $\Z\subset \RR^d$ and let $f: \R \to \Z\subset \RR^d$ be a reward embedding. For $z \in \Z$ let $\tilde{Q}_z$ denote some trained direct representation. Let $r \in \R$  be a bounded reward function and $f(r)=z_r$ its reward embedding. Then the value error of direct methods decomposes into
\begin{equation}
    \Vert Q_r^\ast - \tilde{Q}_{f(r)}\Vert_\infty \leq \inf{\tilde{r} \in f^{-1}(\{z_r\})} \left(\frac{1}{1 - \gamma} \Vert r - \tilde{r} \Vert_\infty + \Vert Q^*_{\tilde{r}} - \tilde{Q}_{z_r}\Vert_\infty \right).
\end{equation}
    
\end{theorem}

\begin{proof}
    The proof is immediate as we can decompose
    \begin{equation}
        \Vert Q_r^\ast - \tilde{Q}_{z_r}\Vert_\infty \leq \Vert Q^*_r - Q^*_{\tilde{r}}\Vert_\infty  + \Vert Q^*_{\tilde{r}} - \tilde{Q}_{z_r}\Vert_\infty
    \end{equation}
    for any $\tilde{r} \in \R$. Restricting the case to only elements in the pre-image of $z_r$ under $f$ combined with Eq. \ref{eq:prop17:touati} completes the proof.
\end{proof}
\newpage

\section{Empirical analysis of the error bounds}
\label{sec:appendix:empirical-error}

\paragraph{Setup.} We instantiate the decomposition of Eq.~\ref{eq::error_decomposition} in a deterministic Four Rooms grid-world ($|\mathcal{S}| = 104$, $|\mathcal{A}| = 4$, $\gamma = 0.95$), small enough that every quantity in Theorems~\ref{thm:1}--\ref{thm:3} is computed rather than estimated: optimal policies, successor features and successor measures are all direct linear solves, and the dynamics are noiseless, so every error we report is attributable to a representation. We sweep three knobs: the granularity of the features, the number of training tasks, and the size of the transition dataset. Each component influences multiple components of the error, as such we cannot not isolate them, however we select specific hyperparameters for example for the sweep over the dataset size such that the influence of the other terms is reduced. For SF-based methods, features $\phi$ are one-hot block indicators, which fixes $\|\phi\|_\infty = 1$, and training task $i$ is the reward $\phi^\top w_i$ exactly, so the assumption that $\pi_i$ is optimal on $\phi^\top w_i$ holds. SF\&GPI needs no function approximator here: its approximation term compares learned SFs against the exact SFs of the \emph{same} policies, an evaluation problem over policies we trained, so a tabular LSTD fit can represent $\psi^{\pi_i}$ exactly and the term reflects only how much of the state-action space the data covered. USF and FB instead need a model of how $\psi$ varies with $w$ (or $z$). We use a bilinear model with one RBF per training task to approximate $\psi(s,a,w)$
\begin{equation*}
\tilde\psi(s,a,w) \;=\; \sum_{k=0}^{n} g_k(w)\,\Theta_k(s,a),
\qquad
g(w) \;=\; \bigl[\,1,\; e^{-\|w - w_1\|^2 / 2\sigma^2},\;\ldots,\; e^{-\|w - w_n\|^2 / 2\sigma^2}\,\bigr].
\end{equation*}
For FB, We fit $B$ in closed form as the top-$d$ eigenvectors of the summed Gram matrix of the training successor measures --- the subspace minimizing $\sum_t |m_t - m_t\Pi_B|_F$.

\paragraph{Measuring the approximation term in value space.} We report
$\varepsilon_{\text{approx}}$ as $\max_i \|(\psi^{\pi_i} - \tilde\psi_i)^\top w_r\|_\infty$,
The reason is that a term comparing representations is not well defined here: in a deterministic grid-world many policies are optimal (i.e. every shortest path) and they share a $Q$ while differing in everything else, so Theorem~\ref{thm:2}'s reference $\psi^{w_r}$ is an arbitrary choice among them. %
The value form is invariant to that tie-breaking, since all policies optimal on $\phi^\top w_r$ satisfy
$\psi^\top w_r = Q^*_{\phi^\top w_r}$.%

\begin{figure}[H]
    \centering
    \includegraphics[width=0.99\linewidth]{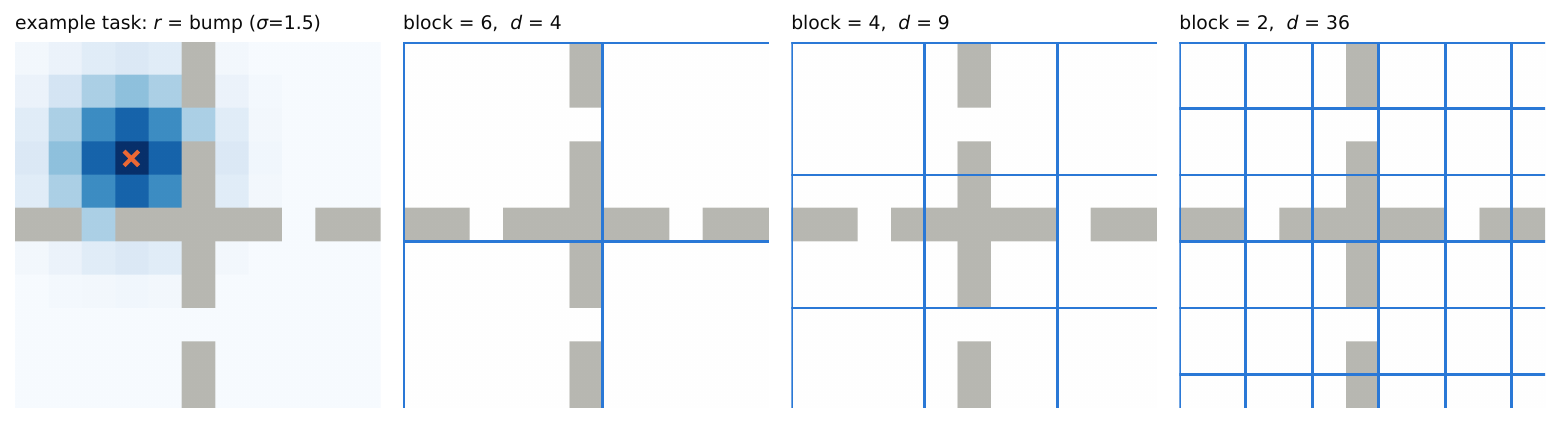}
    \caption{Visualization of the environment used to obtain empirical results on error bounds. Images after the first to the left show the predefined features $\phi$ used for SF-based methods. Rewards are normally distributed around a goal with $\sigma = 1.5$.}
    \label{fig:app_bound_emp_1}
\end{figure}

\begin{figure}[H]
    \centering
    \includegraphics[width=0.99\linewidth]{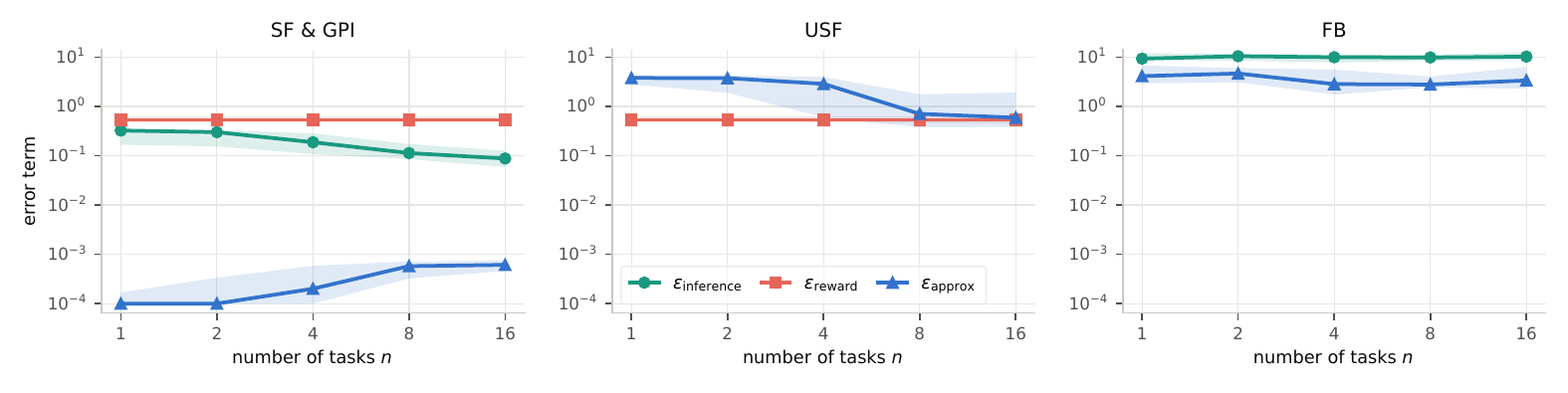}
    \caption{Empirical measurement of error bounds as the number of tasks grows. Results averaged over 50 seeds. For USF and FB this translates in a bigger pool of sampled reward embeddings on which to train. For SF+GPI the successor features of each sampled policy simply are stored. Results are obtained with dataset of size \texttt{D = 10000} and latent dimension \texttt{w = 9} (4x4 blocks).}
    \label{fig:app_bound_emp_2}
\end{figure}

\begin{figure}[H]
    \centering
    \includegraphics[width=0.99\linewidth]{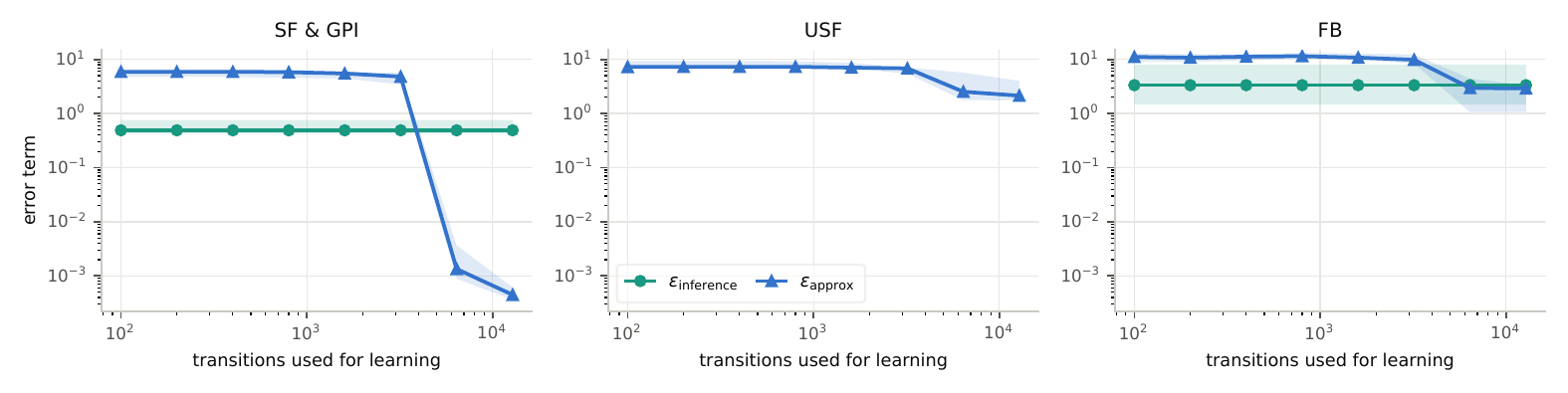}
    \caption{Empirical measurement of error bounds as the number of transitions in the training dataset grows. Results averaged over 50 seeds. Results are obtained with latent dimension \texttt{w = 104} (1x1 blocks) and \texttt{n\_tasks = 16} to isolate the approximation error.}
    \label{fig:app_bound_emp_3}
\end{figure}

\begin{figure}[H]
    \centering
    \includegraphics[width=0.99\linewidth]{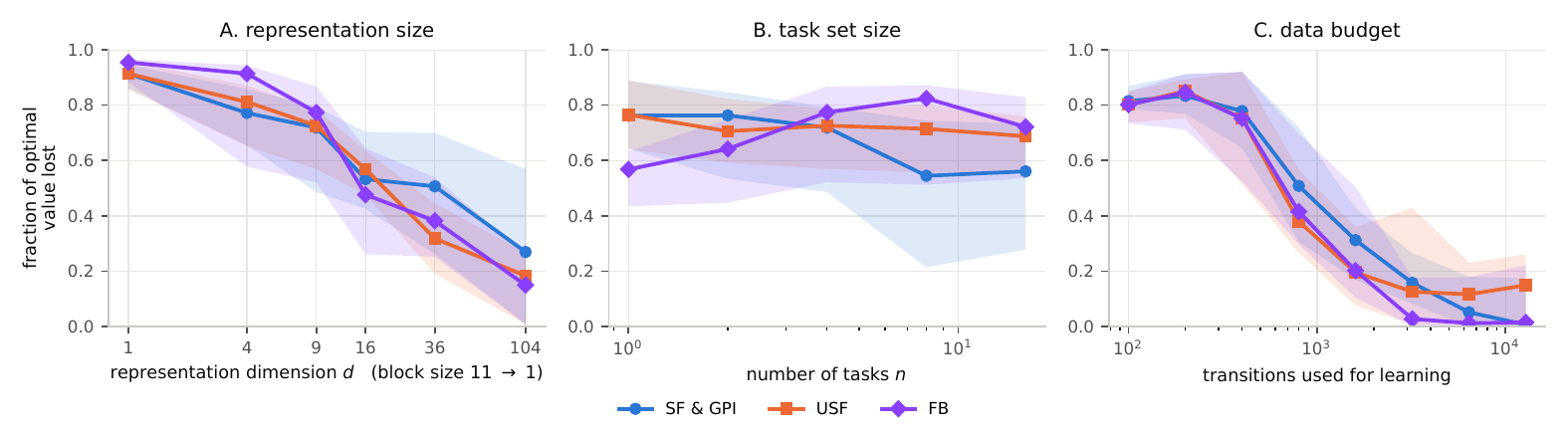}
    \caption{What the terms translate to in practice: the fraction of optimal value the extracted policy gives up, across the same three sweeps. Lines are medians over 50 seeds, bands the inter-quartile range; all methods see identical tasks, and transitions at each setting.}
    \label{fig:app_bound_emp_4}
\end{figure}

\begin{figure}[H]
    \centering
    \includegraphics[width=0.99\linewidth]{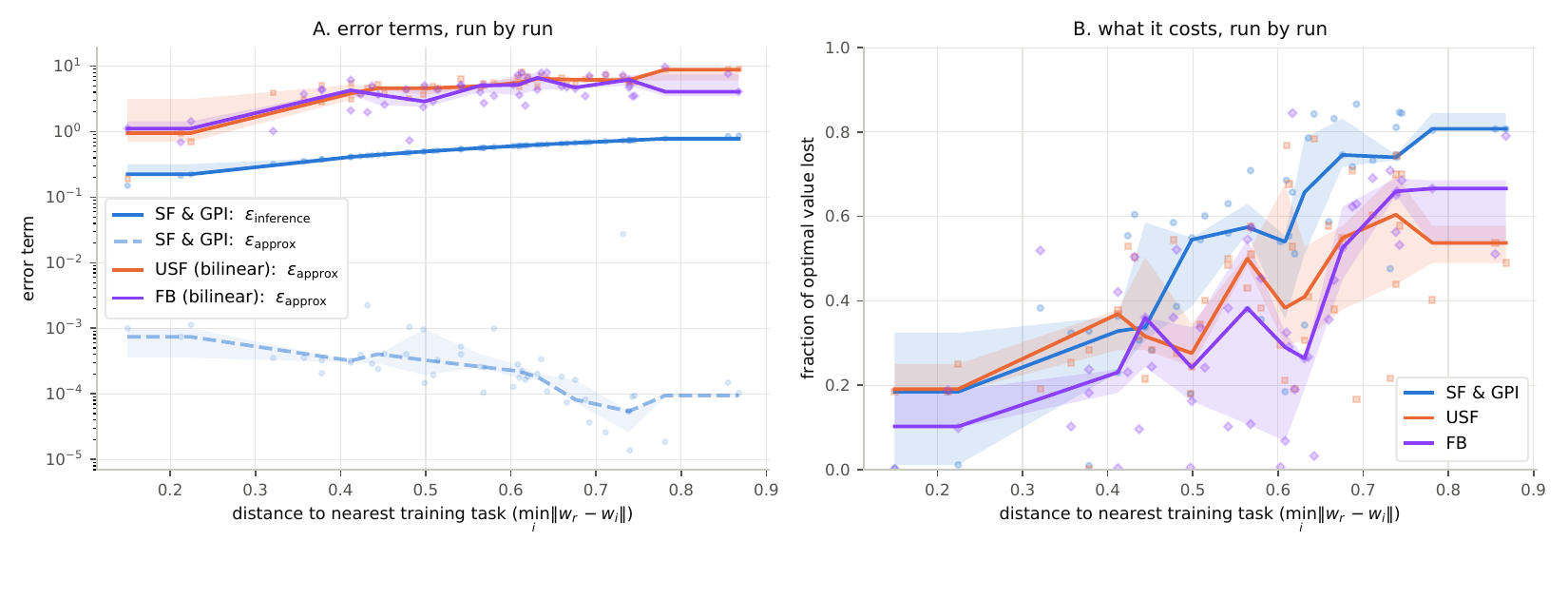}
    \caption{The same measurement with the task set held fixed at \texttt{n\_tasks = 4}, over 50
seeds ordered by the distance from the test task to its nearest training task, $\min_i \|w_r - w_i\|$. Fixing $n$ removes the averaging effect of the codebook sweep and leaves only the per-draw one. \textbf{A}: SF\&GPI books this distance explicitly as $\varepsilon_{\text{inference}}$, while USF and FB, which carry no inference term, show it inside $\varepsilon_{\text{approx}}$ instead. \textbf{B}: the value lost, for the same runs. Lines are per-bin medians, bands the inter-quartile range. Latent dimension \texttt{w = 36} (2x2 blocks), \texttt{D = 10000}.} \label{fig:app_bound_emp_5}
\end{figure}

\newpage
\section{Additional Details Successor Measure}

\noindent The successor measure $M^\pi$ satisfies a Bellman-like equation~\citep{blier2021learning}:
\begin{equation}
        M^{\pi}(s, a, X) 
        = \mathbbm{1}\{(s,a) \in X\}
         + \gamma\,\mathbb{E}_{s' \sim p(\cdot \mid s, a),\, a' \sim \pi(\cdot|s')}
        \left[M^{\pi}(s', a', X)\right].
\end{equation}

\subsection{Parametrized SM loss}
Norm on $M$:
\begin{equation}
    ||M||_\rho^2=\mathbb E_{(s,a)\sim \rho\atop s^+\sim \rho}\Big[\Big(\frac{M(s,a,\mathrm{d}s^+)}{\rho(\mathrm d s^+)}\Big)^2\Big ]
\end{equation}
parametric model of $M$:
\begin{equation}
    m_\theta(s,a,\d s^+) = m_\theta(s,a,s^+)\rho(\d s^+)
\end{equation}
The SM loss is derived by expanding the following TD error:

\begin{align}
    &||M - M^{\text{target}}||_\rho^2 = ||M - (I + \gamma P \bar M)||_\rho^2 \\
    &\hspace{0.2cm}= || m_\theta(s,a,s^+)\rho(\d s^+) - \delta_{(s)}(\d s^+) - \mathbb E_{s'\sim P(\mathrm{d}s'| s,a)\atop a'\sim \pi}[\bar m_\theta(s',a',s^+)\rho(\d s^+)] ||_\rho^2\\
    &\hspace{0.2cm}= \mathbb E_{(s,a)\sim \rho\atop s^+\sim \rho}[(m_\theta(s,a,s^+) - \frac{\delta_{(s)}(\d s^+)}{\rho(\d s^+)} - \mathbb E_{s'\sim P(\mathrm{d}s'| s,a)\atop a'\sim \pi}[\bar m_\theta(s',a',s^+)])^2] \\
    &\hspace{0.2cm}= \mathbb E_{(s,a)\sim \rho\atop s^+\sim \rho}[(\overbrace{m_\theta(s,a,s^+)}^b \underbrace{- \frac{\delta_{(s)}(\d s^+)}{\rho(\d s^+)}}_a \overbrace{- \mathbb E_{s'\sim P(\mathrm{d}s'| s,a)\atop a'\sim \pi}[\bar m_\theta(s',a',s^+)]}^b)^2]
\end{align}
Expand $(a+b)^2$ and move terms independent of $m_\theta$ into $\text{Const}$
\begin{align}
&=-2\mathbb E_{(s,a)\sim \rho\atop s^+\sim \rho}[\underbrace{\frac{\delta_{(s)}(\d s^+)}{\rho(\d s^+)}m_\theta(s,a,s^+)}_{\text{Nonzero only if}\,s\in\d s^+}] + \mathbb E_{(s,a,s')\sim \rho, a'\sim \pi(\cdot|s')\atop s^+\sim \rho }[(m_\theta(s,a,s^+) - \bar m_\theta(s',a',s^+))^2] +\, \text{Const}\\
& =-2\mathbb E_{(s,a)\sim \rho}[m_\theta(s,a,s)] + \mathbb E_{(s,a,s')\sim \rho,a'\sim \pi(\cdot|s')\atop s^+\sim \rho }[(m_\theta(s,a,s^+) - \bar m_\theta(s',a',s^+))^2] +\, \text{Const}
\end{align}

\subsection{FB Loss with Successor Representation 1-shift}
In \cite{touatidoes}, the successor measure, represented as $M^\pi=F^\top B \rho$, is learned by minimizing a slightly different loss to the one in \cite{touati2021learning} which is the one we present in Section~\ref{sec::FB}. This is due to a different definition of the successor measure itself in \cite{touatidoes}, specifically they ignore the immediate occupancy term by multiplying the SM by the transition matrix P. In other words we check if the next state and action are in $X$ rather than the current state and action. Specifically:
\begin{equation}    
\tilde M^\pi(s,a,X) = \EE_{\pi,p}\big[\sum_{t=0}^\infty \gamma^t \mathbbm{1}\{(s_{t+1},a_{t+1})\in X\} \mid s_0=s,a_0=a\big].
\end{equation}
Or in matrix form
\begin{equation}
    \tilde M^\pi = PI + \gamma P_\pi \tilde M
\end{equation}
In words, the successor measure defined as in the above equations ignores the immediate occupancy (i.e. $\mathbbm{1}\{(s_{0},a_{0})\in X\}$). The relationship between the successor measure $M$, as defined in Eq.~(\ref{eq::sm}), and $\tilde M$ is:
\begin{equation}
\tilde M^\pi = P^\pi(I + \gamma P^\pi M)
\end{equation}
Resulting in the following loss for $\tilde M = F^\top B \rho$
\begin{equation}
\label{eq::fb_loss_one_shift}
    \begin{aligned}
        \mathcal{L}(F,B)=\,
        &\mathbb{E}_{(s_t,a_t,s_{t+1})\sim\rho \atop (s',a')\sim \rho}[(F(s_t,a_t,z)^\top B(s'a')-\bar F(s_{t+1},\pi_z(s_{t+1}),z)^\top \bar B(s',a'))^2]\\
        &-2\mathbb{E}_{(s_t,a_t,s_{t+1},a_{t+1})\sim \rho}[F(s_t,a_t,z)^\top B(s_{t+1},a_{t+1})].
    \end{aligned}
\end{equation}

\end{document}